\documentclass[10pt]{article} 
\usepackage[preprint]{tmlr}

\usepackage{amsmath,amsfonts,bm}

\def\eqref#1{equation~\ref{#1}}

\def\1{\bm{1}}

\DeclareMathAlphabet{\mathsfit}{\encodingdefault}{\sfdefault}{m}{sl}
\SetMathAlphabet{\mathsfit}{bold}{\encodingdefault}{\sfdefault}{bx}{n}

\usepackage{hyperref}
\usepackage{url}
\usepackage{tikz}
\usepackage{booktabs}
\usepackage{makecell}
\usepackage{graphicx}
\usepackage{amsmath}
\usepackage{amsfonts}
\usepackage{import}

\title{Learning the Geometry of Admissible Hypotheses through Inductive Bias in Training Distributions}

\author{\name James Crowley \email 	
jamescrowley1857@gmail.com \\
      \addr School of Mechanical and Material Engineering\\
      University College Dublin
      \AND
      \name Faez Ahmed \email faez@mit.edu \\
      \addr Department of Mechanical Engineering\\ Massachusets Institute of Technology
      \AND
      \name Anton van Beek \email anton.vanbeek@ucd.ie \\
      \addr School of Mechanical and Material Engineering\\
      University College Dublin}

\def\month{MM}  
\def\year{YYYY} 
\def\openreview{\url{https://openreview.net/forum?id=XXXX}} 

\begin{document}

\maketitle

\begin{abstract}
Scientific discovery often requires reasoning over competing hypotheses that are consistent with experimental observations. For mixed-variable and combinatorial hypothesis spaces, however, constructing probabilistic representations remains challenging because both the active model components and their associated parameters are unknown. In this work, we present a framework for learning continuous latent representations of admissible partial differential equations (PDEs) by embedding a scientific inductive bias directly into the training distribution. Progressively richer structural principles (e.g., sparsity, logical dependencies, common PDE families, and physical admissibility) are used to generate a structured distribution of hypotheses from which a gated variational autoencoder learns a continuous latent manifold. Experimental results show that the resulting 11-dimensional representation accurately reconstructs a broad collection of representative PDEs, while exhibiting smooth geometric transitions both within and across equation families. Through an ablation study we further demonstrate that introducing scientific principles reduces both structural misclassifications of equation forms and parameter estimation errors when reconstructing a representative benchmark set of admissible partial differential equations. These results show that embedding a scientific inductive bias in the training distribution enables the learning of compact and geometrically meaningful hypothesis manifolds, providing a principled foundation for future inference over competing governing equations.
\end{abstract}
\section{Introduction}
Scientific discovery often requires reasoning over a space of competing hypotheses that are consistent with experimental observations. As these hypothesis spaces grow in size and complexity, manually exploring alternative explanations becomes increasingly difficult, motivating computational methods that can efficiently represent and search candidate scientific models. In this work, we focus on scientific hypotheses that can be expressed as mixed-variable combinatorial objects, where both the active components and their associated continuous parameters are unknown. Examples include molecular structures \cite{boiko2023}, material compositions \cite{merchant2023}, biological regulatory networks \cite{jiao2020}, and governing equations of physical systems \cite{le2018}. Learning probabilistic representations of such hypothesis spaces has the potential to support hypothesis generation, experimental design, and probabilistic scientific inference. However, the mixed-variable, high-dimensional, and highly correlated nature of these spaces makes them challenging to represent with conventional statistical methods.

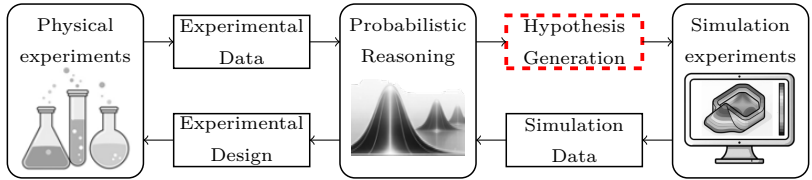
\begin{figure}
\centering
    \begin{tikzpicture}


        \draw[line width=0.2mm, fill = white,rounded corners=2mm] (1mm,2mm) --++ (18mm,0mm) --++ (0mm,23mm) --++ (-18mm,0mm) -- cycle;
        \draw[line width=0.2mm, fill = white,rounded corners=2mm] (45mm,2mm) --++ (18mm,0mm) --++ (0mm,23mm) --++ (-18mm,0mm) -- cycle;
        \draw[line width=0.2mm, fill = white,rounded corners=2mm] (89mm,2mm) --++ (18mm,0mm) --++ (0mm,23mm) --++ (-18mm,0mm) -- cycle;

        \draw[line width=0.2mm, fill = white] (23mm,16.5mm) --++ (18mm,0mm) --++ (0mm,7mm) --++ (-18mm,0mm) -- cycle;
        \draw[line width=0.5mm, fill = white, dashed, draw = red] (67mm,16.5mm) --++ (18mm,0mm) --++ (0mm,7mm) --++ (-18mm,0mm) -- cycle;
        \draw[line width=0.2mm, fill = white] (23mm,3.5mm) --++ (18mm,0mm) --++ (0mm,7mm) --++ (-18mm,0mm) -- cycle;
        \draw[line width=0.2mm, fill = white] (67mm,3.5mm) --++ (18mm,0mm) --++ (0mm,7mm) --++ (-18mm,0mm) -- cycle;
    
        \node[inner sep=0pt] at (10mm,10mm){\includegraphics[width=15mm]{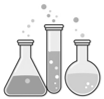}};
        \node[inner sep=0pt] at (54mm,10mm){\includegraphics[width=15mm, height = 10mm]{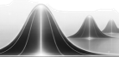}};
        \node[inner sep=0pt] at (98mm,10mm){\includegraphics[width=15mm]{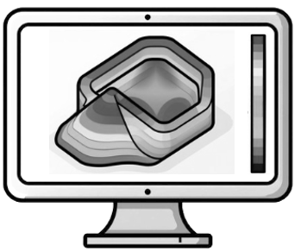}};

        \node[anchor=center, align=center, text width = 18mm] at (32mm,20mm) {\scriptsize Experimental Data};
        \node[anchor=center, align=center, text width = 18mm] at (76mm,20mm) {\scriptsize Hypothesis Generation};

        \node[anchor=center, align=center, text width = 18mm] at (32mm,7mm) {\scriptsize Experimental Design};
        \node[anchor=center, align=center, text width = 18mm] at (76mm,7mm) {\scriptsize Simulation Data};

        \node[anchor=center, align=center, text width = 18mm] at (10mm,20mm) {\scriptsize Physical experiments};
        \node[anchor=center, align=center, text width = 18mm] at (54mm,20mm) {\scriptsize Probabilistic Reasoning};
        \node[anchor=center, align=center, text width = 18mm] at (98mm,20mm) {\scriptsize Simulation experiments};

        \draw[<-] (19mm,7mm) --++ (4mm,0mm);
        \draw[<-] (63mm,7mm) --++ (4mm,0mm);
        \draw[->] (19mm,20mm) --++ (4mm,0mm);
        \draw[->] (63mm,20mm) --++ (4mm,0mm);

        \draw[<-] (41mm,7mm) --++ (4mm,0mm);
        \draw[<-] (85mm,7mm) --++ (4mm,0mm);
        \draw[->] (41mm,20mm) --++ (4mm,0mm);
        \draw[->] (85mm,20mm) --++ (4mm,0mm);



        
        
    \end{tikzpicture}
    \caption{Conceptual framework illustrating the role of probabilistic latent representations in AI-assisted scientific discovery. Physical experiments and numerical simulations jointly inform a learned latent representation of competing scientific hypotheses, which in turn supports predictive modelling, guides future experimentation, and enables probabilistic reasoning over governing equations. The representation of scientific hypotheses is the focus of this work.}
    \label{fig:intro}
\end{figure}

In this work, we consider scientific hypotheses in the form of partial differential equations (PDEs). We envision a mode of AI-assisted scientific discovery in which physical experiments, numerical simulations, and probabilistic latent representations interact to support reasoning over competing governing equations (Figure~\ref{fig:intro}). Within this perspective, the learned latent representation serves as a probabilistic intermediary between physical experiments and simulation models, enabling both sources of evidence to jointly inform posterior beliefs over competing scientific hypotheses. A central challenge is that PDE hypotheses combine discrete decisions regarding which terms are active with continuous unknown coefficients, resulting in a mixed-variable hypothesis space that is both highly structured and strongly correlated \cite{le2018}. Here, we develop a systematic framework for learning latent representations of such hypothesis spaces. Although the integration of physical and simulation data for equation discovery is not new \cite{meidani2024}, existing generative representations are not naturally designed to support inference when the set of competing hypotheses may itself be incomplete \cite{kennedy2001}. Our objective is therefore to construct a probabilistic representation that can support inference over structured governing-equation hypotheses and provide a foundation for future methods that account for model-form inadequacy.

An early approach for equation discovery represents competing scientific hypotheses using a predefined library of candidate equations. In classical system identification, a small collection of physically motivated equation forms is specified a priori, after which the unknown parameters of each candidate are optimized to maximize agreement with experimental observations. The equation with the smallest prediction error is then selected as the governing model \cite{isermann2011}. While effective when suitable candidates are known in advance, this approach relies on prior knowledge of the functional form and therefore provides limited flexibility when discovering previously unknown governing equations \cite{wang2017}.

To reduce this dependence on manually specified models, subsequent methods replaced libraries of complete equations with libraries of candidate function terms from which new combinations can be made through sparse regression \cite{schmidt2009}. Representative examples include AI Feynman for symbolic regression \cite{udrescu2020} and Sparse Identification of Nonlinear Dynamics (SINDy) for discovering governing differential equations from data \cite{le2018,brunton2016,zhang2019,zolman2025}. Rather than selecting from a small set of predefined equations, these methods identify a sparse subset of active terms whose coefficients best explain the observed dynamics. The SINDy framework has subsequently been extended to accommodate moving boundaries \cite{bekar2025}, noisy and sparse measurements \cite{messenger2021}, and latent-space formulations based on autoencoders to improve scalability for high-dimensional systems \cite{conti2023}.

An alternative approach is to incorporate domain knowledge directly into the hypothesis space through mathematical grammars, where production rules define the set of admissible symbolic expressions \cite{chomsky2014}. Such grammars have been used for equation discovery by placing Bayesian priors over grammar production rules, thereby biasing the search toward physically meaningful expressions \cite{brence2021}. Context-free grammars also provide a natural representation for generative modelling because valid expressions can be represented as parse trees and encoded directly using variational autoencoders \cite{kusner2017}. More recently, grammar-constrained variational autoencoders have been proposed to learn continuous latent representations of ordinary differential equation structures, enabling efficient exploration through heuristic optimization and discrete flow models \cite{yu2025,yu2026}. A related alternative is to represent mathematical expressions as directed acyclic graphs, making them amenable to graph-based generative models and graph neural networks \cite{ranasinghe2025}. These approaches demonstrate that incorporating structural prior knowledge can substantially improve the representation and exploration of equation spaces. However, their latent representations primarily describe the symbolic structure of an equation; numerical constants are typically represented as discrete symbolic elements or treated separately from the learned structural representation. Consequently, arbitrary continuous coefficients are not represented jointly with equation structure as part of a single mixed-variable hypothesis. In many equation-discovery settings, the symbolic form is therefore identified first and its continuous coefficients subsequently estimated from data. While effective for identifying governing equations, this separation is less well suited to probabilistic reasoning over complete parameterized hypotheses within a unified latent space.

While existing equation discovery methods have proven successful, they are primarily designed to identify the single hypothesis that best explains a set of observations. In contrast, our objective is to learn a continuous latent representation of scientifically admissible hypotheses, enabling Bayesian inference over multiple competing equations that provide plausible explanations of the observed data. To achieve this, we construct a distribution of training samples using progressively richer domain knowledge, including sparsity, logical dependencies between equation terms, common PDE family structures, and physical admissibility through positive-definite diffusion matrix. Rather than treating these principles as constraints imposed during optimization, we use them to define the distribution of admissible hypotheses from which the model learns. The probabilistic formulation of the variational autoencoder is particularly attractive because it learns a continuous probability distribution over admissible hypotheses \cite{kingma2013}, providing a natural foundation for future Bayesian inference while simultaneously regularizing the latent representation. More broadly, this perspective demonstrates that scientific inductive bias can be introduced through the training distribution itself, rather than through specialized model architectures or additional regularization terms.

Unlike grammar-based latent representations, which encode symbolic equation structures and estimate coefficients in a subsequent optimization stage, our model learns a unified latent representation of complete parameterized partial differential equations. Consequently, equation structure and continuous coefficients are represented jointly within a single probabilistic space. Here, we focus on constructing and analyzing this representation rather than performing Bayesian inference itself \cite{gelman1995}. Specifically, we show that the learned latent representation (i) accurately reconstructs a representative benchmark of governing equations, (ii) exhibits smooth geometric transitions both within and across PDE families, and (iii) organizes physically meaningful equation classes into an interpretable latent geometry. Together, these results demonstrate that the proposed representation accurately captures the structure of the admissible hypothesis space and provides a suitable foundation for future Bayesian inference over both equation structure and continuous parameters. More broadly, we view probabilistic reasoning as a natural framework for scientific discovery because experimental evidence rarely identifies a unique governing hypothesis \cite{fajardo2023}. Instead, multiple competing explanations often remain plausible, motivating representations that support inference over distributions of scientific hypotheses rather than single point estimates.
\section{Representation and Learning of Scientific Hypothesis Spaces: An Application to PDEs}
In this section we introduce a systematic approach toward enforcing an inductive bias to learn hypothesis spaces for PDEs.

\subsection{Overarching Framework}
\begin{figure}
    \begin{tikzpicture}

    \node[anchor=center, align=left, text width = 120mm] at (60mm,80mm) {\scriptsize \textbf{1: Informed data generation through enforcing physics based knowledge}};
    
    \draw[->, line width = 0.5mm] (0mm,75mm) --++ (150mm,0mm);
    \fill[white] (50mm,74mm) --++ (0mm,2mm) --++ (50mm,0mm) --++ (0mm,-2mm) -- cycle;
    \node[anchor=center, align=center, text width = 50mm] at (75mm,75mm) {\scriptsize \textbf{Increasing Inductive Bias}};

    \fill[white!20!gray] (0mm,63mm) --++ (0mm,8mm) --++ (30mm,0mm) --++ (0mm,-8mm) -- cycle;
    \fill[white!37!gray] (30mm,63mm) --++ (0mm,8mm) --++ (30mm,0mm) --++ (0mm,-8mm) -- cycle;
    \fill[white!55!gray] (60mm,63mm) --++ (0mm,8mm) --++ (30mm,0mm) --++ (0mm,-8mm) -- cycle;
    \fill[white!73!gray] (90mm,63mm) --++ (0mm,8mm) --++ (30mm,0mm) --++ (0mm,-8mm) -- cycle;
    \fill[white!90!gray] (120mm,63mm) --++ (0mm,8mm) --++ (30mm,0mm) --++ (0mm,-8mm) -- cycle;
    \node[anchor=center, align=center, text width = 30mm] at (15mm,67mm) {\scriptsize \textbf{None}};
    \node[anchor=center, align=center, text width = 30mm] at (45mm,67mm) {\scriptsize \textbf{Sparsity}};
    \node[anchor=center, align=center, text width = 30mm] at (75mm,67mm) {\scriptsize \textbf{Logical \\ (dependencies)}};
    \node[anchor=center, align=center, text width = 30mm] at (105mm,67mm) {\scriptsize \textbf{Classification} \\ (PDE families)};
    \node[anchor=center, align=center, text width = 30mm] at (135mm,67mm) {\scriptsize \textbf{Physics} \\ (positive-definiteness)};

    \foreach \i in {0,...,4}{
        \draw[black!50!white, line width = 0.5mm] (\i * 30mm, 35mm) --++ (0mm,36mm) --++ (30mm,0mm) --++ (0mm,-36mm) -- cycle;
    }
    
    \node[anchor=center, align=center, text width = 30mm] at (15mm,40mm) {\scriptsize Arbitrary combinations of terms};
        \foreach \i in {1,...,15}{
        \pgfmathsetmacro{\posX}{rnd * 2.6 + 0.2}
        \pgfmathsetmacro{\posY}{rnd * 1.6 + 4.5}
        \pgfmathsetmacro{\radius}{0.1}
        \filldraw[black, fill=black!20!white] 
            (\posX,\posY) circle (\radius cm);
    }

    \node[anchor=center, align=center, text width = 30mm] at (45mm,40mm) {\scriptsize Limit number of active terms};
    \foreach \i in {1,...,3}{
        \pgfmathsetmacro{\posX}{3 + 0.35*\i}
        \pgfmathsetmacro{\posY}{5.4}
        \pgfmathsetmacro{\radius}{0.1}
        \filldraw[black, fill=black!20!white] 
            (\posX,\posY) circle (\radius cm);
    }
    \foreach \i in {4,...,6}{
        \pgfmathsetmacro{\posX}{3 + 0.35*\i}
        \pgfmathsetmacro{\posY}{5.4}
        \pgfmathsetmacro{\radius}{0.1}
        \filldraw[black, fill=white] 
            (\posX,\posY) circle (\radius cm);
    }
    \node[anchor=center, align=center, text width = 30mm] at (55mm,54mm) {\scriptsize \( \cdots \)};
    \node[anchor=center, align=center, text width = 30mm] at (45mm,48mm) {\scriptsize (e.g., \( \leq 6\) active terms)};

    \node[anchor=center, align=center, text width = 30mm] at (75mm,40mm) {\scriptsize Physically consistent term combinations};
    \filldraw[black, fill = white!55!gray] (62mm,56mm) --++ (0mm,4mm) --++ (6mm,0mm) --++ (0mm,-4mm) -- cycle;
    \filldraw[black, fill = white!55!gray] (62mm,46mm) --++ (0mm,4mm) --++ (6mm,0mm) --++ (0mm,-4mm) -- cycle;
    \filldraw[black, fill = white!55!gray] (77mm,56mm) --++ (0mm,4mm) --++ (6mm,0mm) --++ (0mm,-4mm) -- cycle;
    \filldraw[black, fill = white!55!gray] (72mm,46mm) --++ (0mm,4mm) --++ (6mm,0mm) --++ (0mm,-4mm) -- cycle;
    \filldraw[black, fill = white!55!gray] (82mm,46mm) --++ (0mm,4mm) --++ (6mm,0mm) --++ (0mm,-4mm) -- cycle;

    \draw[->, line width = 0.25mm] (65mm,56mm) --++ (0mm,-6mm);
    \draw[->, line width = 0.25mm] (80mm,56mm) --++ (-5mm,-6mm);
    \draw[->, line width = 0.25mm] (80mm,56mm) --++ (5mm,-6mm);
    \node[anchor=center, align=center, text width = 30mm] at (65mm,58mm) {\scriptsize \(uu_x\)};
    \node[anchor=center, align=center, text width = 30mm] at (65mm,48mm) {\scriptsize \(u_x\)};
    \node[anchor=center, align=center, text width = 30mm] at (80mm,58mm) {\scriptsize \(u_{xy}\)};
    \node[anchor=center, align=center, text width = 30mm] at (75mm,48mm) {\scriptsize \(u_{xx}\)};
    \node[anchor=center, align=center, text width = 30mm] at (85mm,48mm) {\scriptsize \(u_{yy}\)};

    \node[anchor=center, align=center, text width = 30mm] at (105mm,40mm) {\scriptsize Equations from common PDE families};
    \filldraw[black, fill=blue] (95mm,59mm) circle (1mm);
    \filldraw[black, fill =orange] (94mm,54mm) --++ (0mm,2mm) --++ (2mm,0mm) --++ (0mm,-2mm) -- cycle;
    \filldraw[black, fill = green] (94mm,50mm) --++ (1mm,2mm) --++ (1mm,-2mm) -- cycle;
    \filldraw[black, fill = purple] (95mm,46mm) --++ (1mm,1mm) --++ (-1mm,1mm) --++ (-1mm,-1mm) -- cycle;

    \node[anchor=center, align=left, text width = 22mm] at (108mm,59mm) {\scriptsize : Elliptic};
    \node[anchor=center, align=left, text width = 22mm] at (108mm,55mm) {\scriptsize : Parabolic};
    \node[anchor=center, align=left, text width = 22mm] at (108mm,51mm) {\scriptsize : Hyperbolic};
    \node[anchor=center, align=left, text width = 22mm] at (108mm,47mm) {\scriptsize : First-order \(\cdots \)};

    \node[anchor=center, align=center, text width = 30mm] at (135mm,40mm) {\scriptsize Physically valid equations};

    \node[anchor=center, align=center, text width = 30mm] at (135mm,52mm) {\scriptsize \(  \begin{aligned} \mathbf{A} & = \left[\begin{matrix}
        a_{11} & a_{12} & a_{13} \\
        a_{21} & a_{22} & a_{23} \\
        a_{31} & a_{32} & a_{33} 
    \end{matrix}\right]\\
    &\lambda_i(\mathbf{A})\geq 0, \quad \forall i \end{aligned} \)};

    \filldraw[black, fill = white!90!gray] (0.5mm,18mm) --++ (0mm,4mm) --++ (13mm,0mm) --++ (0mm,-4mm) -- cycle;
    \filldraw[black, fill = white!90!gray] (15mm,18mm) --++ (0mm,4mm) --++ (32mm,0mm) --++ (0mm,-4mm) -- cycle;
    \filldraw[black, fill = white!90!gray] (0.5mm,13mm) --++ (0mm,4mm) --++ (13mm,0mm) --++ (0mm,-4mm) -- cycle;
    \filldraw[black, fill = white!90!gray] (15mm,13mm) --++ (0mm,4mm) --++ (32mm,0mm) --++ (0mm,-4mm) -- cycle;
    \filldraw[black, fill = white!90!gray] (0.5mm,8mm) --++ (0mm,4mm) --++ (16mm,0mm) --++ (0mm,-4mm) -- cycle;
    \filldraw[black, fill = white!90!gray] (18mm,8mm) --++ (0mm,4mm) --++ (29mm,0mm) --++ (0mm,-4mm) -- cycle;

    \node[anchor=center, align=left, text width = 50mm] at (25mm,30mm) {\scriptsize \textbf{2: Structured dataset}};
    \node[anchor=center, align=left, text width = 48mm] at (25mm,20mm) {\scriptsize \(u_t=u_{xx}\quad u_t =u_{xx}+u_{yy}+u_{zz}\)};
    \node[anchor=center, align=left, text width = 48mm] at (25mm,15mm) {\scriptsize \(u_x=u_{xx} \quad u_x=0.5u_{xx}+0.2u_{yy}\)};
    \node[anchor=center, align=left, text width = 48mm] at (25mm,10mm) {\scriptsize \( 0=u_{x}+u_{y} \quad 0=0.1u_{xx}+0.9u_{yy}\)};
    \node[anchor=center, align=center, text width = 48mm] at (25mm,5mm) {\scriptsize \( \cdots\)};
    
    \node[anchor=center, align=left, text width = 60mm] at (80mm,30mm) {\scriptsize \textbf{3: Manifold learning}};

    \filldraw[black, fill = white!90!gray ,rounded corners=1mm] (56mm,5mm) --++ (0mm,20mm) --++ (7mm,-3mm) --++ (0mm,-14mm) -- cycle;
    \filldraw[black, fill = white!90!gray ,rounded corners=1mm] (97mm,8mm) --++ (0mm,14mm) --++ (7mm,3mm) --++ (0mm,-20mm) -- cycle;

    \draw[->, line width = 0.25mm,rounded corners=1mm] (48mm,22mm) --++ (2mm,0mm) --++ (0mm,-7mm) --++ (6mm,0mm);
    \draw[->, line width = 0.25mm,rounded corners=1mm] (48mm,5mm) --++ (2mm,0mm) --++ (0mm,10mm) --++ (6mm,0mm);
    \draw[->, line width = 0.25mm,rounded corners=1mm] (104mm,15mm) --++ (6mm,0mm);
    \node[anchor=center, align=center, text width = 48mm] at (53mm,17mm) {\scriptsize \( {\mathbf{X}}\)};
    \node[anchor=center, align=center, text width = 48mm] at (108mm,17mm) {\scriptsize \( \hat{\mathbf{X}}\)};
    \node[anchor=center, align=center, text width = 48mm, rotate = 90] at (60mm,15mm) {\scriptsize \( q_{\Phi}(\mathbf{z}|\mathbf{X})\)};
    \node[anchor=center, align=center, text width = 48mm, rotate = 90] at (100mm,15mm) {\scriptsize \( p_{\Theta}(\mathbf{X}|\mathbf{z})\)};
    
    \foreach \i in {1,...,35}{
        \pgfmathsetmacro{\posX}{rnd * 1 + 7}
        \pgfmathsetmacro{\posY}{rnd * 0.7 + 1.9}
        \pgfmathsetmacro{\radius}{0.1}
        \filldraw[black, fill=blue] 
            (\posX,\posY) circle (\radius cm);
    }
    \foreach \i in {1,...,35}{
        \pgfmathsetmacro{\posX}{rnd * 0.7 + 8}
        \pgfmathsetmacro{\posY}{rnd * 1 + 1.6}
        \pgfmathsetmacro{\radius}{0.03}
        \filldraw[black, fill=orange] 
            (\posX,\posY) --++ (0mm,2mm) --++ (2mm,0mm) --++ (0mm,-2mm) -- cycle;
    }
        \foreach \i in {1,...,35}{
        \pgfmathsetmacro{\posX}{rnd * 1.5 + 7.2}
        \pgfmathsetmacro{\posY}{rnd * 1.2 + 0.5}
        \pgfmathsetmacro{\radius}{0.03}
        \filldraw[black, fill=green] 
            (\posX,\posY) --++ (1mm,2mm) --++ (1mm,-2mm) -- cycle;
    }
        \foreach \i in {1,...,35}{
        \pgfmathsetmacro{\posX}{rnd * 0.5 + 6.7}
        \pgfmathsetmacro{\posY}{rnd * 0.5 + 1.2}
        \pgfmathsetmacro{\radius}{0.03}
        \filldraw[black, fill=purple] 
            (\posX,\posY) --++ (1mm,1mm) --++ (-1mm,1mm) --++ (-1mm,-1mm) -- cycle;
    }

    \node[anchor=center, align=left, text width = 40mm] at (130mm,30mm) {\scriptsize \textbf{4: Use of latent space}};
    \node[inner sep=0pt] at (130mm,15mm){\includegraphics[width=25mm]{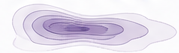}};
    \node[anchor=center, align=left, text width = 40mm] at (130mm,25mm) {\scriptsize Experimental observations: \(\mathbf{Y}\)};
    \node[anchor=center, align=left, text width = 40mm] at (130mm,19mm) {\scriptsize Posterior over \( \mathbf{z}: P(\textbf{z}|\textbf{Y})\)};

    \node[anchor=center, align=left, text width = 40mm] at (130mm,8mm) {\scriptsize \( u_{t}=u_{xx}\)};
    \node[anchor=center, align=left, text width = 40mm] at (130mm,5mm) {\scriptsize \( u_{x}=u_{xx}\)};
    \node[anchor=center, align=left, text width = 40mm] at (130mm,2mm) {\scriptsize \( \cdots \)};

    \filldraw[black, fill = red!50!blue] (123mm,7mm) --++ (0mm,2mm) --++ (22mm,0mm) --++ (0mm,-2mm) -- cycle;
    \filldraw[black, fill = red!50!blue] (123mm,4mm) --++ (0mm,2mm) --++ (11mm,0mm) --++ (0mm,-2mm) -- cycle;
    \filldraw[black, fill = red!50!blue] (123mm,1mm) --++ (0mm,2mm) --++ (3mm,0mm) --++ (0mm,-2mm) -- cycle;

    \draw[->, line width = 0.25mm,rounded corners=1mm] (130mm,22mm) --++ (0mm,-2mm);
    \draw[->, line width = 0.25mm,rounded corners=1mm] (130mm,11.5mm) --++ (0mm,-2mm);

    \end{tikzpicture}
    \caption{Flowchart of the proposed approach for systematically adding an increasing restrictive inductive bias to establish a structured data set of physically admissible equations. Subsequently, such a dataset can be used to learn a low-dimensional continuous manifold that can be used to establish posterior distributions over training data contain simulation and physical experimental data to test which equations most likely.}
    \label{fig:flowchart}
\end{figure}
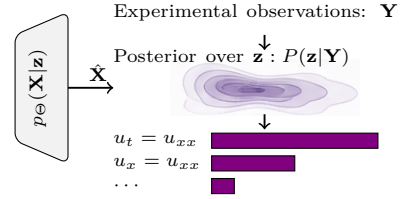

The central idea of the proposed framework is that scientific knowledge can be used to introduce statistical regularity into a hypothesis space before training a machine learning model. Specifically, rather than relying on the learning algorithm to discover structure from an arbitrary collection of hypotheses, we first construct a training distribution whose organization reflects established scientific principles. Figure~\ref{fig:flowchart} provides an overview of this process. The upper row illustrates how progressively stronger forms of scientific knowledge are incorporated to transform an initially unstructured hypothesis space into one exhibiting meaningful regularity. Without such structure, the space of partial differential equations consists of arbitrary combinations of active terms and continuous coefficients, resulting in a high-dimensional mixed-variable combinatorial space that is difficult to represent and explore. The objective is therefore not to restrict the search to a small collection of predefined equations, but rather to construct a tractable hypothesis space that remains sufficiently expressive while reflecting established scientific knowledge.

The regularity introduced into the training distribution is motivated by physical principles governing PDEs. Similar in spirit to mathematical grammars, these principles define which hypotheses are considered scientifically admissible. However, rather than explicitly specifying production rules for every symbolic expression, we formulate a small collection of high-level design principles that collectively induce structure in the hypothesis space. While this perspective is conceptually related to grammar-based representations, the role of scientific knowledge is fundamentally different. Rather than explicitly enumerating symbolic production rules that define admissible expressions, we use scientific principles to shape the probability distribution from which hypotheses are sampled. Consequently, the inductive bias is learned statistically through the latent representation rather than enforced symbolically during generation. In our framework, we progressively introduce four levels of inductive bias: (i) \textit{sparsity}, by limiting the number of active terms; (ii) \textit{logical consistency}, by enforcing dependencies between related terms; (iii) \textit{equation family consistency}, by restricting equations to common classes of PDEs; and (iv) \textit{physical admissibility}, by requiring diffusion operators to satisfy positive-definiteness. Notably, the final criterion depends jointly on both the symbolic structure of an equation and its continuous coefficients, illustrating the importance of representing both components within a unified hypothesis space. While these principles are motivated by PDEs, they are not intended to represent the only possible choice. Instead, they demonstrate how domain knowledge can systematically be embedded into the training distribution through progressively richer inductive bias.

Once these principles have been established, a structured training dataset of admissible hypotheses can be generated, as illustrated in the lower-left portion of Figure~\ref{fig:flowchart}. This dataset is subsequently used to train a variational autoencoder, which learns a continuous latent representation of the structured hypothesis space. Because the training distribution already reflects scientific regularities, the learned manifold organizes related equations in nearby regions of the latent space while preserving smooth transitions between different hypothesis classes. The resulting latent representation provides a compact geometric description of scientifically admissible hypotheses that forms the basis for the subsequent analyses presented in this work and, more broadly, establishes a suitable foundation for future Bayesian inference over competing scientific models.

\subsection{Equation Representation and Inductive Bias}
We begin by considering a general partial differential equation of the form 
\begin{equation}
    \quad F\left(\nabla^ku,\nabla^{k-1}u,\ldots,\nabla u,u,x \right)=0,
\end{equation}
where \(\nabla\) is a differential operator, \(k\) is the maximum order, and \(u\) is a function \(u: U\rightarrow \mathbb{R}\) of the variables \(\mathbf{x}=\left\{x_1,\ldots x_n\right\}\in U\subset\mathbb{R}^n\). Although this representation is continuous, practical equation discovery requires selecting a finite collection of candidate differential operators together with their associated coefficients. Consequently, hypothesis generation becomes a mixed-variable combinatorial problem consisting of both discrete structural decisions and continuous parameters. Specifically, this space would be defined as \(\mathbb{R}^{n^k}\times \mathbb{R}^{n^{k-1}}\times,\ldots,\times\mathbb{R}^n\times\mathbb{R}\times U\) that is dependent on the maximum order of the equation considered \(k\). Here we restrict ourselves to second order equations (i.e., \(k=2\)). 

under this construction, we would like to establish a training dataset \(\mathcal{D}=\left\{\mathbf{X}_1,\ldots,\mathbf{X}_m\right\}\) with \(m\) different realizations of admissible PDEs. The individual PDEs in this case are defined through a finite \(p\)-dimensional vector \(\mathbf{X}_i=\left\{X_{i,1},\ldots,X_{i,p} \right\}\). Finally, we can divide the vectors of admissible PDEs into two components \(\mathbf{X}_i=\left\{ \mathbf{b}_i, \mathbf{c}_i \right\} \) where \(\mathbf{b}_i\in\left\{0,1 \right\}^{p/2}\) are binary variables that indicate if a term is active, and \( \mathbf{c}_i\in \mathbb{R}^{p/2}\) are the continuous coefficients. This gives us the foundation to construct the space of admissible hypothesis as
\begin{equation}
    \mathcal{X}=\left\{(\mathbf{b}, \mathbf{c}):\mathbf{b}\in\left\{0,1 \right\}^{p/2},\mathbf{c}\in \mathbb{R}^{p/2} \right\}.
\end{equation}
Our objective is to learn a mapping \(f:\mathcal{X}_{phy}\rightarrow \mathcal{Z}\) that embeds admissible hypotheses into a continuous latent space. For this purpose, we would like to construct progressively smaller subsets of the form
\begin{equation}
\mathcal{X}\supset \mathcal{X}_{spr}\supset \mathcal{X}_{log}\supset \mathcal{X}_{fam}\supset \mathcal{X}_{phy},
\end{equation}
where \(\mathcal{X}_{spr}\) is the hypothesis space where only sparsity principles are enforces, \( \mathcal{X}_{log}\) is the set where we also enforce logical dependencies, \( \mathcal{X}_{fam}\) is the set where we also enforce that equations should belong to common family structures, and finally \(\mathcal{X}_{phy}\) is the set that also enforces physical principles. The first three principles progressively constrain the symbolic structure of admissible hypotheses, whereas the fourth additionally constrains their continuous parameterization.

\begin{itemize}
    \item \textbf{Principle 1 - Sparsity:} Here we make the assumption that physics-based equations typically have a relatively small number of active terms. We define this as
    \begin{equation}
        \mathcal{X}_{spr}=\left\{(\mathbf{b},\mathbf{c})\in \mathcal{X}:  k_{min}\leq ||\mathbf{b}||_0\leq k_{max} \right\},
    \end{equation}
    where \( ||\cdot||_0\) is the cardinality of a vector, and \(  k_{min},  k_{max}\) are the minimum and maximum number of active terms. This assumption reflects the observation that many governing equations are dominated by a relatively small number of active terms, rather than implying that all physical systems are inherently sparse. Here we use \(  k_{min}=2\) and \( k_{max}=6\).
    \item \textbf{Principle 2 - Logical Dependencies:} Next, we consider that PDEs typically following structural principles (e.g., some terms are commonly only observed to be active if others are active). Such information can be used to define the set of PDEs that satisfies a set logical dependencies as
    \begin{equation}
        \mathcal{X}_{log}=\left\{\mathbf{x}\in \mathcal{X}_{spr}: \psi(\mathbf{b})=1 \right\},
    \end{equation}
    where \(\psi(\mathbf{b}) \) is a dependency operator that equals one when satisfied. We consider two categories of logical dependencies.
    \begin{itemize}
        \item \textbf{Gradient dependencies} that ensure that nonlinear transport terms are only introduced if the underlying linear differential operators exists. Examples of this include,
    \begin{equation}
    uu_x\Rightarrow u_x, \quad uu_y\Rightarrow u_y, \quad u_x^2\Rightarrow u_x, \quad u_y^2\Rightarrow u_y.
    \end{equation}
        \item \textbf{Diffusion dependencies} to ensure that mixed diffusion terms have an associated principle direction. Example of this include,
    \begin{equation}
        u_{xy}\Rightarrow\left\{u_{xx},u_{yy} \right\}, \quad u_{xz}\Rightarrow\left\{u_{xx},u_{zz} \right\}, \quad u_{yz}\Rightarrow\left\{u_{yy},u_{zz} \right\}.
    \end{equation}
    \end{itemize}
    We do have to emphasize here that these Logical dependencies encode structural relationships between differential operators that commonly occur in governing equations, and not that it is fundamentally impossible to for example have a PDE containing \(uu_x\) but not \( u_x \). Their purpose is to eliminate combinations of terms that are mathematically inconsistent with the classes of PDEs considered in this work, while retaining a sufficiently broad hypothesis space. 
    \item \textbf{Principle 3 - PDE Family Classification:} Next, we assume that each of the PDEs has to belong a commonly encountered family of equations that collective represent a broad space of admissible equations. Specifically, 
    \begin{equation}
        \mathcal{X}_{fam}=\left\{\mathbf{x}\in \mathcal{X}_{spr}: \phi(\mathbf{x})\in\mathcal{F} \right\},
    \end{equation}
    where \(\mathcal{F}=\{\text{elliptic, parabolic, hyperbolic, steady first order, time dependent first order\}}\) is the set of PDE families considered in this work. To classify these families, we can resort to the description of a second-order PDE in operator form, that is defined as
    \begin{equation}
        \mathcal{A} u_{tt}+ \mathcal{B} u_t+\mathbf{v}^T\nabla u =\nabla(\mathbf{A}_s\nabla u),
    \end{equation}
    where \(\mathcal{A}\) governs the second-order time dynamics, \(\mathcal{B}\) governs first-order time dynamics, \(\mathbf{v}\) is the transport (advection) vector, and \(\mathbf{A}_s\) is the symmetric diffusion matrix \cite{evans2022}. Under this description the five families of equations differ only in which of these quantities are present. With that in mind;
    \begin{itemize} 
    \item \textbf{Steady First-order PDEs:} Only first-order transport terms are presents. 
    \item \textbf{Time Dependent First-order PDEs:} First-order transport and first order temporal derivatives are presents. 
    \item \textbf{Elliptic:} Diffusion is present while temporal derivatives are absent.
    \item \textbf{Parabolic:} Diffusion and first-order temporal derivatives are present.
    \item \textbf{Hyperbolic:} Diffusion and second-order temporal derivatives are present, and the second second-order temporal derivative is opposite in sign from the diffusion terms.
    \end{itemize}
    A summary of these PDE family properties has been provided in Table~\ref{tbl:PDE_family_classification}.Similar as with Principle 2, while PDEs certainly can exist that do not fall under the families consider in this work. Our focus is to create regularity without trying to constrict the manifold of learned hypotheses too much.  
    \item \textbf{Principle 4 - Physical Admissibility:} Rather than only considering the diffusion matrix \( \mathbf{A}_s\) is presents we can realize that it is directly informed through the continuous variables \(\mathbf{c}\) (i.e., \(\mathbf{A}_s(\mathbf{c})\)). The diffusion matrix is made up of a set of constants that in order to be physically admissible require to be positive definite (i.e., \( \lambda_i\left(\mathbf{A}_s\right)>0, \quad \forall_i\), where \( \lambda_i\) is the \(i^{th}\) eigen value). As such, the space of physically admissible equations can be defined as
    \begin{equation}
        \mathcal{X}_{phy}=\left\{\mathbf{x}\in \mathcal{X}_{fam}: \lambda_{min}\left(\mathbf{A}_s\right)>0 \right\}, 
    \end{equation}
    where \(\lambda_{min}\left(\mathbf{A}_s\right)\) is the smallest eigen value of the diffusion matrix. It should be noted that the first three principles constrain the symbolic structure of admissible equations. The fourth additionally constrains their parameterization.
\end{itemize}

\begin{table}
\setlength{\tabcolsep}{2.0pt}
\renewcommand{\arraystretch}{1}

\caption{Summary of classification of PDEs that belong to common families, that includes elliptic, parabolic, hyperbolic, steady first order, time dependent first order.}
\centering
{\footnotesize 
\begin{tabular}{llll}
\toprule
 PDE Family & Time terms (i.e., \(u_t\), \(u_{tt}\)) & Transport \(\mathbf{v}^T\nabla u\) & Diffusion \(\mathbf{A}_s\)\\
\midrule
Steady First-order & No & Yes & No\\
Time Dependent First-order & \( u_t\) & Yes & No \\
Elliptic & No & Optional & Yes \\ 
Parabolic & \( u_t \) & Optional & Yes \\
Hyperbolic & \( u_{tt} \) & Optional & Yes \\
\bottomrule 
\end{tabular}
}
\label{tbl:PDE_family_classification}
\end{table}

\subsection{Learning the Hypothesis Manifold}
The structured training dataset is used to learn a continuous latent representation through a gated variational autoencoder (GVAE). Unlike conventional VAEs, which assume a single observation model, the GVAE accommodates mixed-variable inputs by combining Gaussian likelihoods for continuous coefficients with Bernoulli likelihoods for binary activation variables \cite{ma2020}. The GVAE provides the probabilistic machinery required to embed mixed-variable hypotheses into a continuous latent space. Importantly, however, the geometric organization of this space is not imposed by the network architecture itself but emerges from the structured training distribution introduced in the previous subsection.

After applying the structural principles introduced in the previous subsection, the resulting representation consisted of 56 binary activation variables and 11 continuous coefficients. While every candidate term is associated with a binary activation variable, not every active term requires an independent continuous parameter because several coefficients are fixed through normalization or become functionally dependent through the logical and physical constraints imposed on the hypothesis space. Approximately \(120,000\) admissible PDEs were generated for training. To prevent the learned manifold from being dominated by either very sparse or highly populated equations, the number of active linear operators was sampled from a prescribed distribution. Specifically, PDEs containing two, three, four, five, six, and seven active linear terms were sampled with probabilities \(\left\{12, 30, 28, 18, 8, 4 \right\}\%\), respectively. This distribution reflects the observation that the majority of physically meaningful equations considered in this work contain approximately three to four active operators, while still retaining sufficient diversity to learn richer structures.

The encoder and decoder employed symmetric fully connected architectures with hidden layer widths of \(256\), \(128\), and \(64\) neurons. Models were trained for \(200\) epochs using the Adam optimizer with a learning rate of \(1\times10^{-3}\) and a batch size of \(64\). To encourage stable optimization while avoiding posterior collapse, the weight of the Kullback-Leibler (KL) regularization term annealed using a half-cosine wave from 0 to 0.1 during the first 80 training epochs.
\section{Geometric Analysis of Latent Representations}
The objective of the proposed framework is not merely to reconstruct individual equations, but to learn a continuous hypothesis space that represents the geometry of scientifically admissible PDEs. Therefore, in this section we explore four complementary questions: (i) whether the latent representation accurately reconstructs representative PDEs, (ii) whether the learned manifold captures the diversity of admissible hypotheses, (iii) how progressively introducing scientific principles into the training distribution influences the learned representation, and (iv) whether the resulting latent space exhibits smooth and interpretable organization within and across PDE families.

\subsection{Emergence of Latent Geometry}
\label{ssec:latent_dim}
\begin{figure}
    \begin{tikzpicture}

        \node[inner sep=0pt] at (73mm,75mm){\includegraphics[height=48mm]{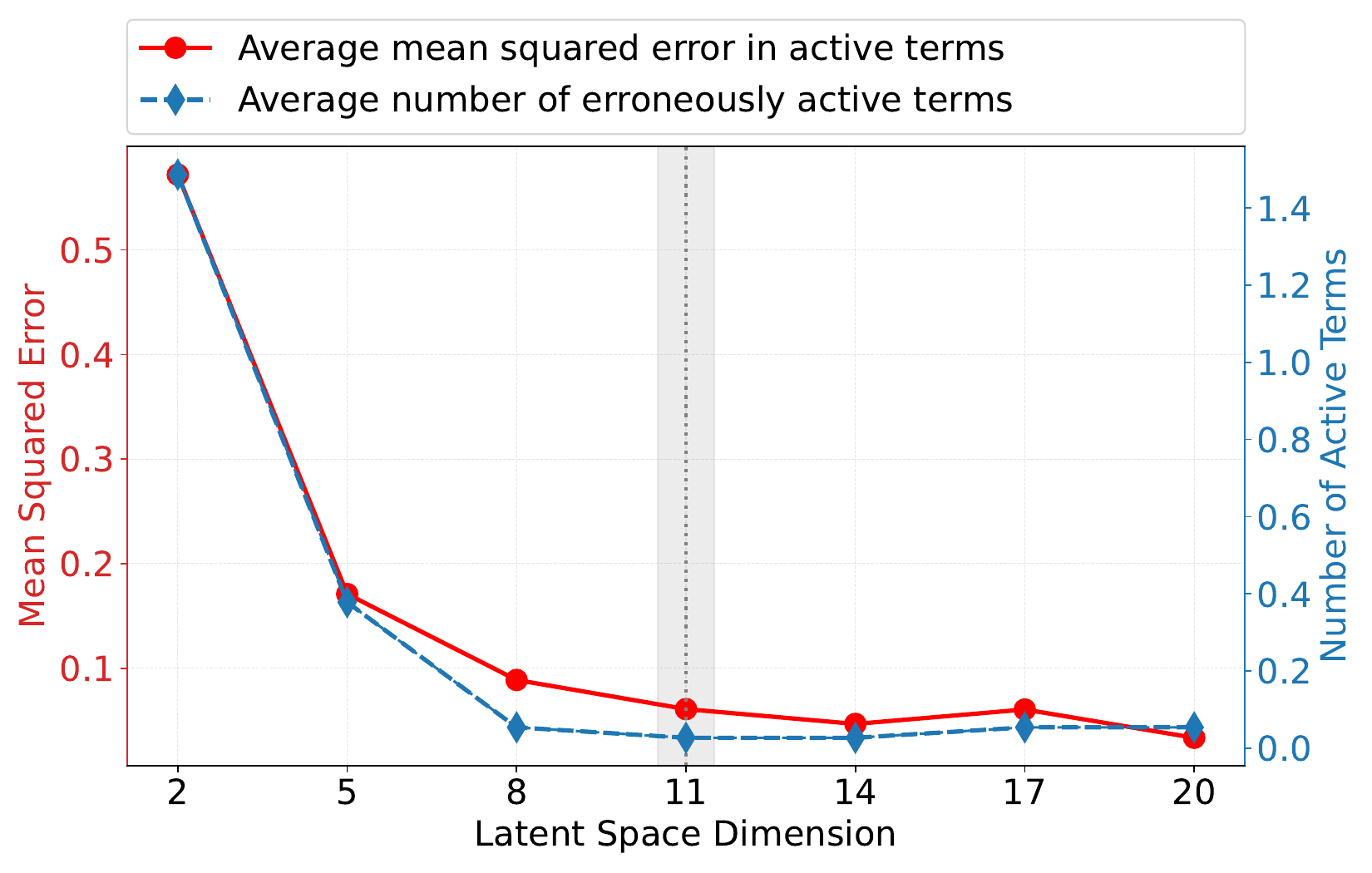}};
        \node[inner sep=0pt] at (73mm,25mm){\includegraphics[height=48mm]{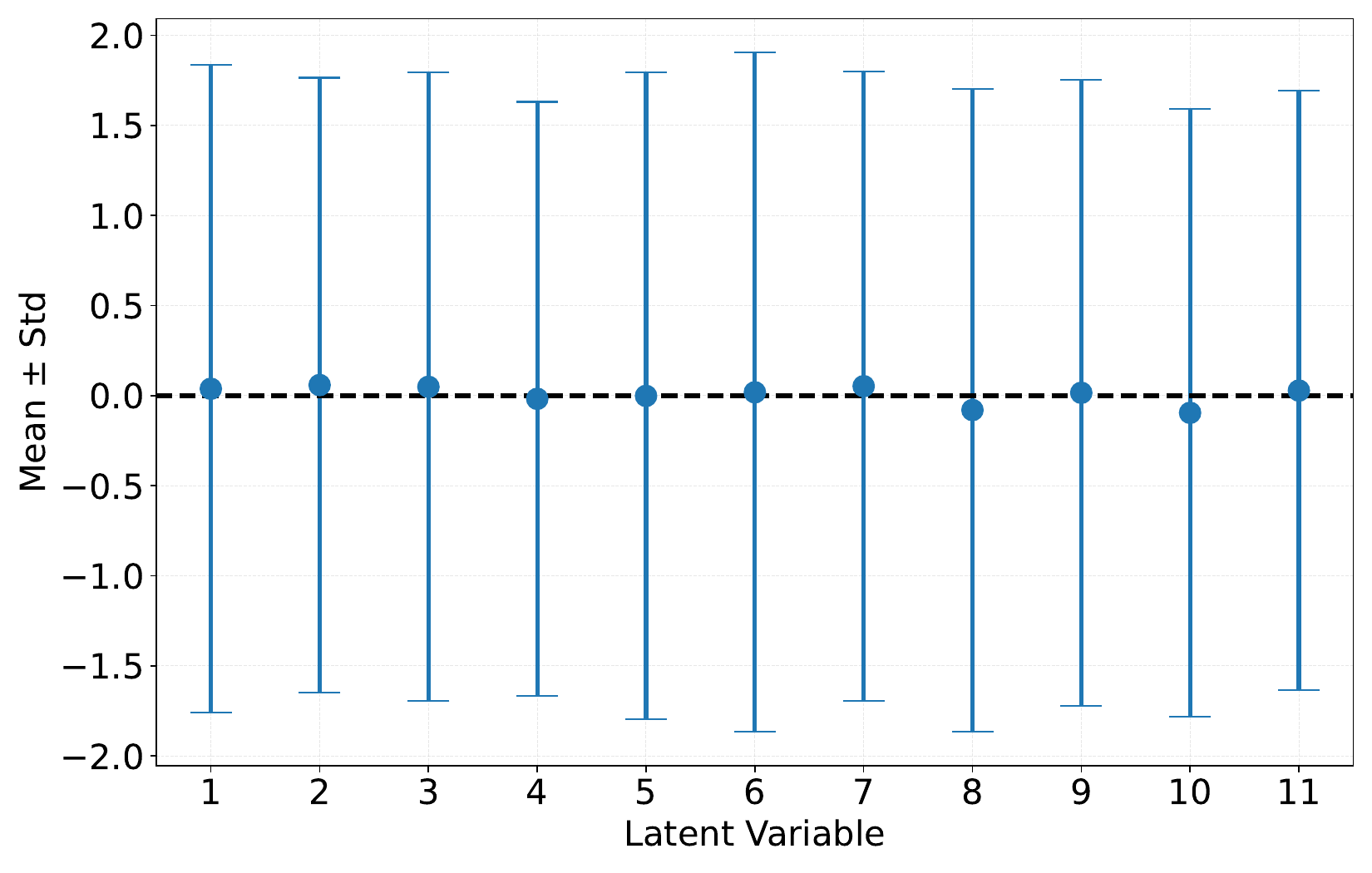}};

        \node[inner sep=0pt] at (-7mm,25mm){\includegraphics[height=48mm,trim=2cm 0cm 0cm 0cm]{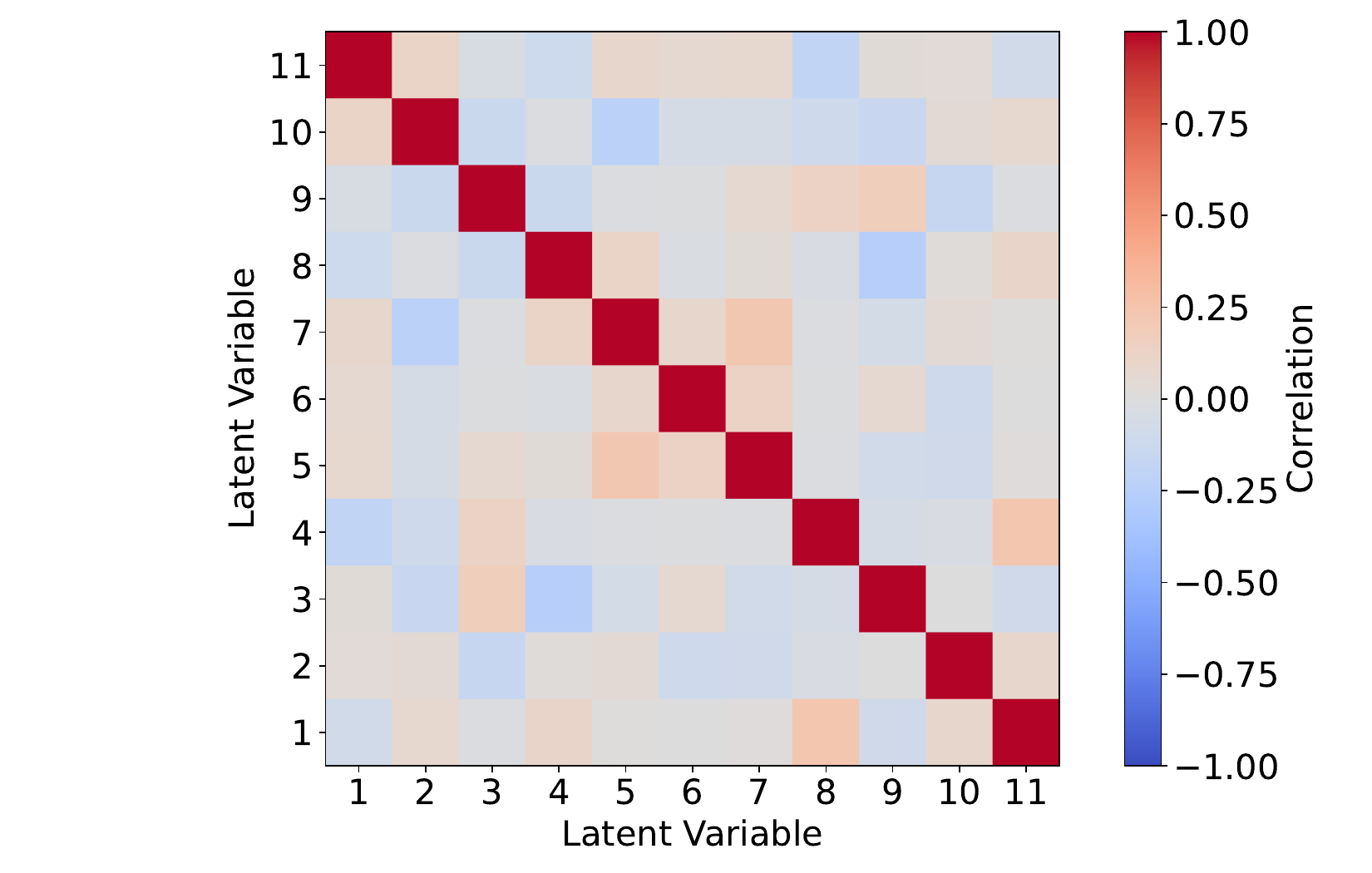}};
        \node[inner sep=0pt] at (-2mm,75mm){\includegraphics[height=48mm]{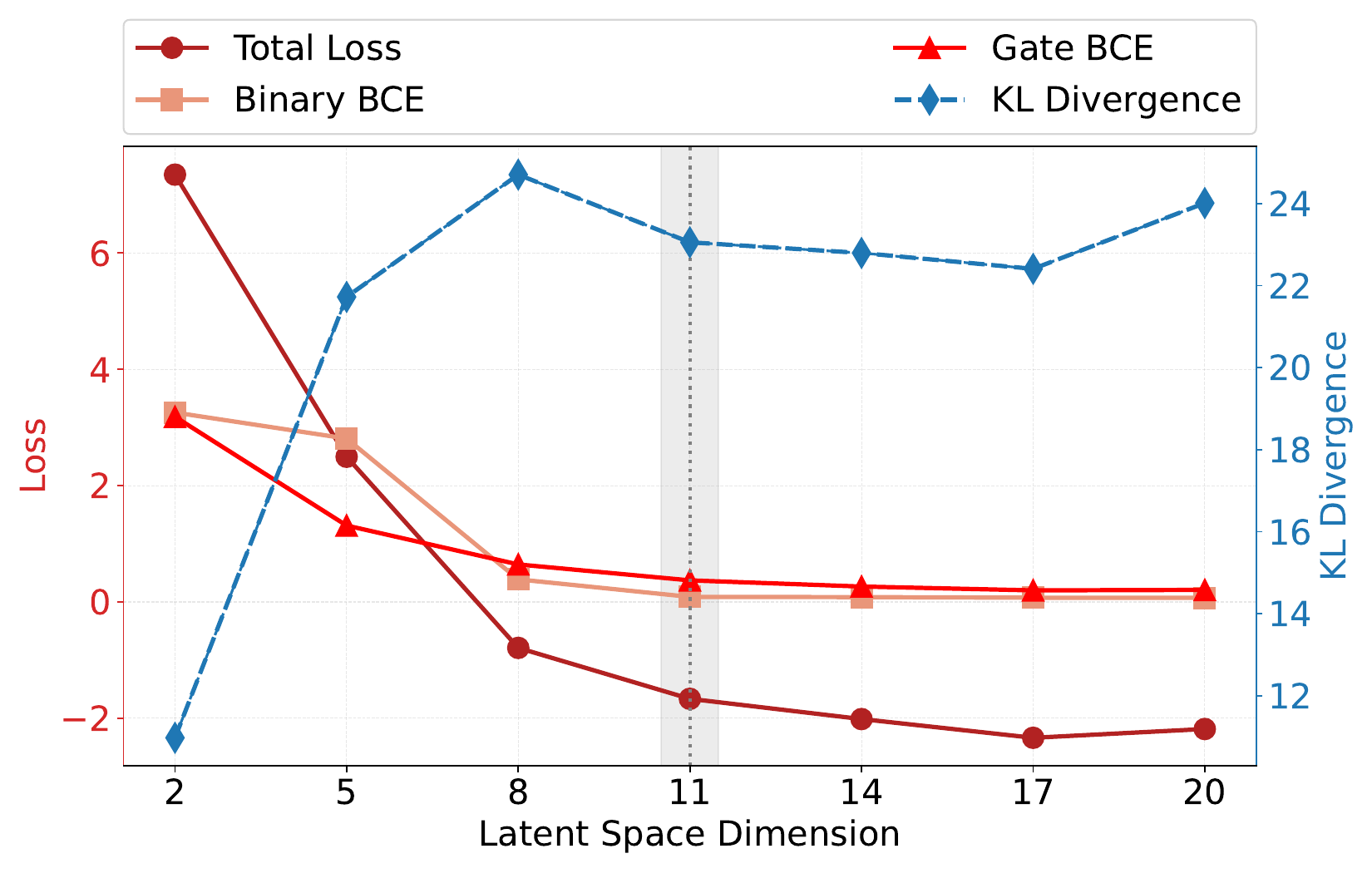}};

        \node[inner sep=0pt] at (-6mm,-25mm){\includegraphics[height=48mm]{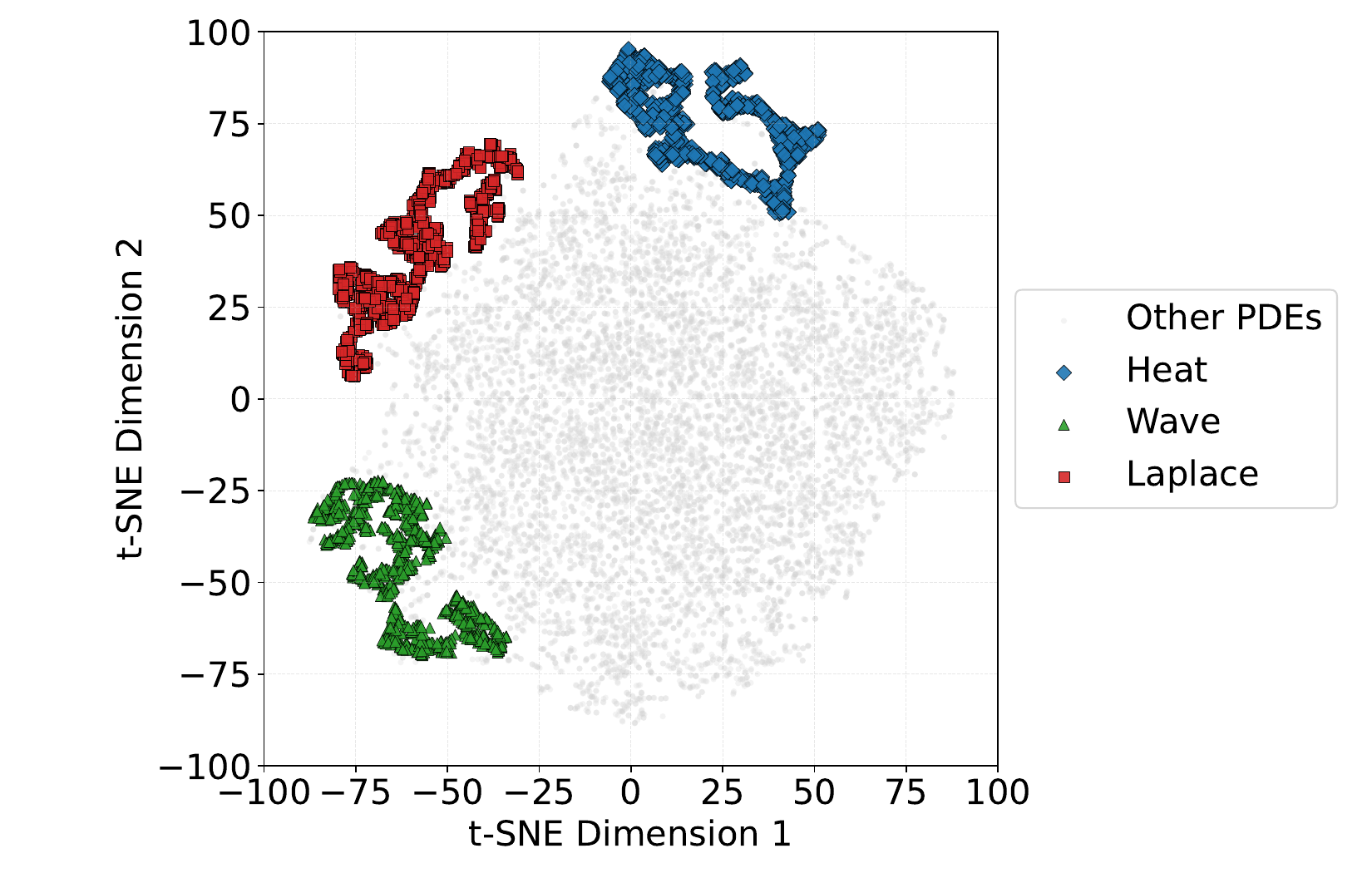}};

        \draw[line width=0.2mm, fill = white] (-40mm,96mm) --++ (4mm,0mm) --++ (0mm,4mm) --++ (-4mm,0mm) -- cycle;
        \node[anchor=center, align=center, text width = 5mm] at (-38mm,98mm) {\scriptsize \textbf{A}};

        \draw[line width=0.2mm, fill = white] (36mm,96mm) --++ (4mm,0mm) --++ (0mm,4mm) --++ (-4mm,0mm) -- cycle;
        \node[anchor=center, align=center, text width = 5mm] at (38mm,98mm) {\scriptsize \textbf{B}};

        \draw[line width=0.2mm, fill = white] (-40mm,46mm) --++ (4mm,0mm) --++ (0mm,4mm) --++ (-4mm,0mm) -- cycle;
        \node[anchor=center, align=center, text width = 5mm] at (-38mm,48mm) {\scriptsize \textbf{C}};

        \draw[line width=0.2mm, fill = white] (36mm,46mm) --++ (4mm,0mm) --++ (0mm,4mm) --++ (-4mm,0mm) -- cycle;
        \node[anchor=center, align=center, text width = 5mm] at (38mm,48mm) {\scriptsize \textbf{D}};

        \draw[line width=0.2mm, fill = white] (-40mm,-4mm) --++ (4mm,0mm) --++ (0mm,4mm) --++ (-4mm,0mm) -- cycle;
        \node[anchor=center, align=center, text width = 5mm] at (-38mm,-2mm) {\scriptsize \textbf{E}};
    \end{tikzpicture}
    \caption{Training results for different latent space dimensions. (A) Breakdown of the training loss into the binary cross-entropy for continuous-variable activation (Gate BCE), binary-term activation (Binary BCE), total loss, and Kullback-Leibler (KL) divergence. The shaded region indicates the latent dimension selected for all subsequent experiments. (B) Minimum error and number of erroneously active terms averaged of the set of 37 benchmark equations in Table~\ref{tbl:canonical_inc}. (C) Correlation matrix of the learned latent variables. (D) Mean and standard deviation of each latent variable over 10,000 randomly generated PDEs, showing approximately zero-centered latent coordinates with consistent variance across dimensions. (E) t-SNE projection of the latent representations highlighting heat, Laplace, and wave equations, demonstrating distinct organization of these equation families within the learned manifold. Collectively, these results indicate that an 11-dimensional latent space provides sufficient capacity to accurately represent the admissible PDE space while maintaining a structured and statistically well-behaved latent representation.}
    \label{fig:dim_analysis}
\end{figure}
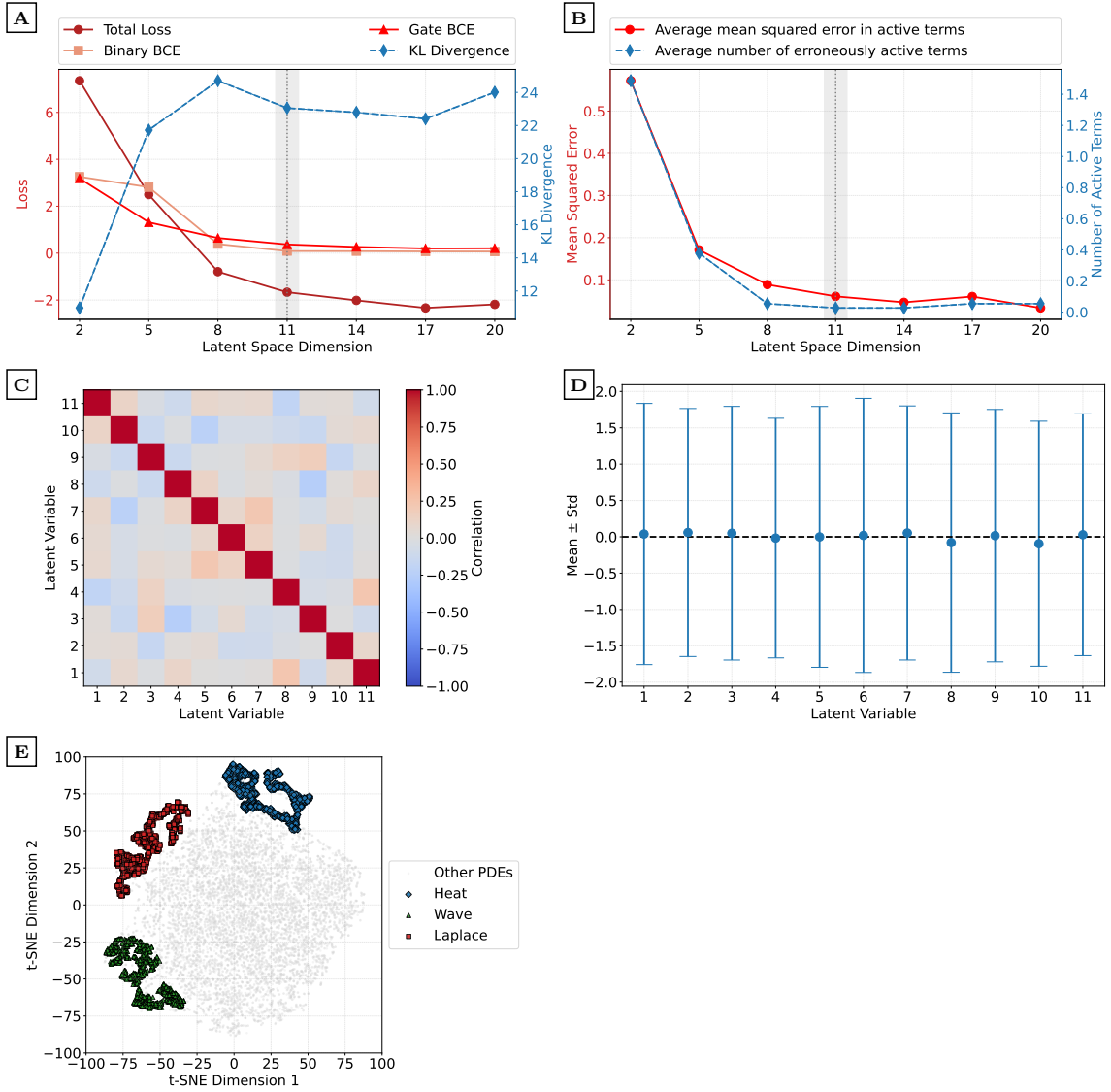

A natural question is how many latent variables are required to represent the space of admissible hypotheses. To investigate this, we trained the GVAE using latent dimensions of \(||\mathbf{z}||_0= \left\{2,5,8,11,14,17,20 \right\}\). The corresponding training losses are shown in Panel A of Figure~\ref{fig:dim_analysis}. As expected, the total training loss decreases as the latent dimensionality increases. Moreover, the binary and continuous reconstruction losses, represented through the binary cross entropcy (BCE), decrease rapidly up to approximately 11 latent variables, after which further improvements become marginal. In contrast, the KL divergence continues to improve for larger latent spaces, indicating that additional latent variables primarily increase conformity of the learned latent distribution to the multivariate Gaussian prior rather than improving reconstruction accuracy. This suggests that the intrinsic dimensionality of the hypothesis space is approximately 11, beyond which additional latent variables provide diminishing practical benefit. Consequently, all subsequent experiments use an 11-dimensional latent representation.

A subsequent question is whether the learned latent space provides sufficient coverage of the admissible hypothesis space defined by the established inductive bias. To assess this, we selected 37 PDEs that we consider to span the hypothesis space induced by our four principles. Subsequently, we attempted to recover each equation through a multi-start optimization in the latent space using 25 random initializations. The target and reconstructed equations are reported in Columns 3 and 4 of Table~\ref{tbl:canonical_inc}, together with the corresponding mean squared errors (MSEs). We find that 34 of the 37 equations are recovered with MSEs around 0.1, demonstrating that the learned latent representation accurately captures the majority of the benchmark equations considered.

\begin{table}
\setlength{\tabcolsep}{2.0pt}
\renewcommand{\arraystretch}{1}

\caption{List of 37 benchmark PDEs and their most accurate reconstructions found in the learned latent space. The results demonstrate that most meaningful PDEs can be found in the learned latent space despite the need to enforce restrictions on the space of admissible forms.}
\centering
{\footnotesize 
\begin{tabular}{l l p{3.6cm} p{6.0cm} r}
\toprule
 ID & PDE Name & Target PDE & Reconstructed PDE & MSE \\
\midrule
\multicolumn{5}{l}{First-Order, Transport, and Nonlinear Transport}\\
\midrule
1 & 1D Advection (+) & \(-u_t+u_x\)  & \(-1.001u_t+0.9998u_x\) &  0.072 \\
2 & 2D Advection & \(-u_t+u_x+u_y \) & \( -0.9987u_t+1.0001u_x+0.9993u_y \) & 0.093  \\
3 & 3D Advection & \( -u_t+u_x+u_y+u_z \) & \( -1.001u_t+1.000u_x+0.9994u_y + 1.0004u_z \) & 0.062 \\
4 & 1D Advection (-) & \( u_t + u_x \) & \( 1.0018u_t + -0.9990u_x \) & 0.139 \\
5 & Steady-State 2D Transport & \( u_x+u_y \) & \( 0.9998u_x+0.9999u_y \) &  0.014\\
6 & 2D Transport & \( -u_t+u_x+0.5 u_y\) & \( -1.000u_t+0.9995u_x+0.4998u_y\) & 0.032\\
7 & Steady-State 3D Transport & \( u_x+u_y+u_z\)& \(0.9997 u_x+0.9991u_y+0.9997u_z\) & 0.051\\
8 & Nonlinear Transport & \( -u_t+u_x+uu_x\)& \( -1.0009u_t+0.9996u_x+uu_x\) & 0.042\\
9 & 1D Hamilton-Jacobi & \( -u_t+u_x+u_xu_x\) & \( -1.0050u_t+0.9999u_x+u_xu_x\) & 0.168\\
10 & 2D Hamilton-Jacobi & \( -u_t+u_x+u_y+u_xu_x \) & \( -0.9989u_t+0.9988u_x+0.9997u_y+u_xu_x \) & 0.065\\
\midrule
\multicolumn{5}{l}{Parabolic} \\
\midrule
11& 1D Heat & \( -u_t+u_{xx} \) & \( -1.0008u_t+0.9994u_{xx} \) & 0.069\\
12& 2D Heat & \( -u_t+u_{xx}+u_{yy}\) & \( -0.9996u_t+0.9989u_{xx}+0.9988u_{yy}\) & 0.090\\
13& 3D Heat & \(-u_t+u_{xx}+u_{yy}+u_{zz} \)& \(-0.9956u_t+1.0013u_{xx}+1.0004u_{yy}+0.9971u_{zz} \) & 0.225\\
14& Anisotropic 2D Heat& \( -u_t+u_{xx}+0.5u_{yy} \) & \( -1.0004u_t+0.9996u_{xx}+0.5002u_{yy} \) & 0.031 \\
15& Anisotropic 3D Heat& \( -u_t+u_{xx}+0.5u_{yy}+0.7u_{zz} \) & \( -0.9997u_t+0.9992u_{xx}+0.5001u_{yy}+0.6998u_{zz} \) & 0.035\\
16& 3D Diffusion \& Reaction & \( -u_t+u_{xx}+u_{yy}+u_{zz}+u \) & \( -1.0005u_t+1.0003u_{xx}+1.0004u_{yy}+1.0000u_{zz}+u \) & 0.025\\
17& 1D Convection Diffusion (+)& \( -u_t+u_x+u_{xx} \) & \( -1.0008u_t+0.9999u_x+1.0002u_{xx} \) & 0.035\\
18& 1D Convection Diffusion (-)& \( -u_t-u_x+u_{xx} \) &  \( -1.0014u_t-0.9999u_x+1.0002u_{xx} \) & 0.056\\
19& 2D Convection Diffusion & \( -u_t+u_x+u_{y}+u_{xx}+u_{yy} \) & \( -0.9949u_t+1.0001u_x+0.9995u_{y}+0.9984u_{xx}+0.9983u_{yy} \) & 0.200\\
20& 3D Convection Diffusion & \( -u_t-u_x-u_{y}-u_{z}+u_{xx}+u_{yy}+u_{zz} \) &  \( -1.0072u_t-0.9997u_x+0.9990u_{y}-0.9990u_{z}+0.9995u_{xx}+0.9967u_{yy}+0.9922u_{zz} \) &  0.301\\
21& 1D Diffusion \& Reaction & \( -u_t+u_{xx}+u\) & \( -1.0019u_t+0.9996u_{xx}+u\)  & 0.078 \\
22& Nonlinear Reaction-Diffusion & \( -u_t+u_{xx}+u_{yy}+u_{zz}+u+uu\) & \( -1.0030u_t+0.9989u_{xx}+0.9998u_{yy}+0.9996u_{zz}+u+uu\) & 0.078\\
23& 1D Burgers' Equation & \( -u_t+uu_x+u_{xx} \) & \( -1.0000u_t+uu_x+1.0017u_{xx} \)& 0.057\\
24& 2D Burgers' Equation & \( -u_t+uu_x+uu_y+u_{xx}+u_{yy} \)& \( -0.9958u_t+1.0062u_{xx}+0.9545u_{yy} \) & 41.119\\
25& 3D Burgers' Equation & \( -u_t+uu_x+uu_y+uu_z+u_{xx}+u_{yy}+u_{zz} \) & \( -1.0012u_t+1.0246u_{xx}+1.0187u_{yy}+0.9774u_{zz} \) & 43.817 \\
\midrule
\multicolumn{5}{l}{Elliptic} \\
\midrule
26& 1D Laplace & \(u_{xx}\) &  \(1.0000u_{xx}\) & 0.004 \\
27& 2D Laplace (v1) & \(u_{xx}+u_{yy}\) & \(0.9993u_{xx}+0.9984u_{yy}\) & 0.114\\
28& 2D Laplace (v2) & \(u_{xx}+u_{zz}\) & \(1.0000u_{xx}+0.9992u_{zz}\) & 0.043\\
29& 3D Laplace & \(u_{xx}+u_{yy}+u_{zz}\) & \(1.0017u_{xx}+0.9978u_{yy}+0.9964u_{zz}\) & 0.250 \\
30& 3D Anisotropic Laplace & \(u_{xx}+0.5u_{yy}+0.25u_{zz}\) & \(1.0011u_{xx}+0.5003u_{yy}+0.2509u_{zz}\) & 0.077\\
31& Helmholtz Equation & \( u_{xx}+u_{yy}+u_{zz}+u\) & \( 0.9991u_{xx}+0.9988u_{yy}+0.9870u_{zz}+u\)  & 0.379\\
32& Linear Second Order Elliptic & \( u_{xx}+u_{yy}+u_{zz}+0.05u_{xy}\) & \( 0.9998u_{xx}+0.9961u_{yy}+1.0006u_{zz}+0.0440u_{xy}\) & 0.266\\
\midrule
\multicolumn{5}{l}{Hyperbolic} \\
\midrule
33& 1D Wave & \(-u_{tt}+u_{xx}\) & \(-0.9998u_{tt}+0.9998u_{xx}\)  & 0.023\\
34& 2D Wave & \(-u_{tt}+u_{xx}+u_{yy}\)  &  \(-0.9984u_{tt}+0.9994u_{xx}+0.9986u_{yy}\) & 0.118\\
35& 3D Wave & \(-u_{tt}+u_{xx}+u_{yy}+u_{zz}\)& \(-0.9991u_{tt}+1.0006u_{xx}+1.0002u_{yy}+0.9983u_{zz}\) & 0.083\\
36& Klein-Gordon & \(-u_{tt}+u_{xx}+u_{yy}+u_{zz}+u\) & \(-0.9938u_{tt}+0.9986u_{xx}+1.0002u_{yy}+0.9964u_{zz}+u\)  & 0.228 \\
37& Telegraph Equation & \(-u_{tt}-u_t+u_{xx}+u_{yy}+u_{zz}\) & \(0.9704u_{xx}+1.0279u_{yy}+0.9521u_{zz}\) & 42.108\\
\bottomrule
\end{tabular}
}
\label{tbl:canonical_inc}
\end{table}

The two- and three-dimensional Burgers' equations fail to recover their nonlinear transport terms because Principle 2 requires nonlinear operators to appear together with their corresponding linear differential operators. Consequently, these equation structures are absent from the training distribution and cannot be represented by the learned latent space. Similarly, the Telegraph equation cannot be reconstructed because Principle 3 excludes PDEs containing both first- and second-order temporal derivatives, preventing this equation family from appearing during training. Interestingly, the one-dimensional Burgers' equation is recovered accurately, indicating that the limitation is not the nonlinear term itself but rather the particular structural assumptions imposed on higher-dimensional variants. These results demonstrate that the learned latent space accurately represents the hypothesis space induced by the proposed training distribution. Moreover, they illustrate that the inductive bias not only improves learning efficiency, but also determines which scientific hypotheses can ultimately be represented and explored. Careful design of these principles is therefore essential, as they define the scope of subsequent inference.

Additional evidence supporting the choice of an 11-dimensional latent space is provided in Panel B of Figure~\ref{fig:dim_analysis}, which reports the average reconstruction error over the 37 benchmark PDEs together with the average number of erroneously active terms. Both metrics decrease rapidly with increasing latent dimensionality before plateauing at approximately 11 latent variables, providing further evidence that this dimension is sufficient to capture the dominant structure of the admissible hypothesis space.

Beyond reconstruction accuracy, a desirable property of the learned representation is that the latent variables exhibit approximately independent Gaussian behavior. The reason that this is desirable is that approximately independent Gaussian latent variables provide a convenient prior distribution from which posterior distributions over the hypothesis space can subsequently be inferred. Panel C of Figure~\ref{fig:dim_analysis} therefore includes a plot of the correlation matrix computed from \(10,000\) randomly generated PDEs. The learned latent variables show relatively weak pairwise correlations, with the majority lying between \(-0.25\) and \(0.25\), indicating that the GVAE has largely disentangled the dominant sources of variation in the hypothesis space. Panel D further summarizes the marginal distributions of the latent variables. While the latent means remain close to zero, the standard deviations are consistently larger than the unit variance imposed by the Gaussian prior, typically lying between \(1.5\) and \(1.9\). This reflects the trade-off between latent regularization and reconstruction accuracy. Specifically, the KL divergence encourages a standard normal latent distribution, the optimization permits moderate deviations when they improve reconstruction of admissible PDEs. Importantly, the standard deviations remain relatively consistent across latent dimensions, suggesting that no single latent variable dominates the learned representation. Collectively, these results indicate that the GVAE learns a well-structured and approximately disentangled latent space that is suitable for probabilistic sampling and subsequent Bayesian inference.

Finally, we investigate whether the learned latent representation exhibits meaningful geometric organization. Panel E of Figure~\ref{fig:dim_analysis} presents a two-dimensional t-SNE projection of the learned latent space \cite{vandermaaten2008}, illustrating parameterizations of the Heat, Wave, and Laplace equations using blue diamonds, green triangles, and red squares, respectively. Although the projection serves only as a visualization of the high-dimensional latent space, it reveals three well-separated clusters corresponding to the different equation families. This organization is further supported by the distances between the latent centroids of each family. The Heat and Laplace equations have the smallest centroid separation (1.669), followed by the Laplace and Wave equations (2.870), while the Heat and Wave equations are separated by the largest distance (4.518). This ordering is physically meaningful. The Heat and Laplace equations both describe diffusion processes and differ primarily through the presence of a first-order temporal derivative, whereas the Wave equation introduces second-order temporal dynamics and therefore belongs to a fundamentally different class of governing equations. The learned latent representation therefore captures relationships that reflect the underlying mathematical structure of the governing equations, suggesting that distances in the latent space provide a meaningful measure of similarity between competing scientific hypotheses.

\subsection{Ablation Study of Scientific Inductive Bias}
Having established that the learned latent representation captures meaningful geometric structure, we next investigate the contribution of the proposed scientific inductive bias. Specifically, we ask whether progressively introducing structural regularity into the training distribution improves the quality of the learned hypothesis space. Unlike conventional ablation studies that remove architectural components, we instead remove progressively richer scientific principles while keeping the GVAE architecture fixed. This isolates the contribution of the training distribution itself.

\begin{figure}
    \begin{center}
    \begin{tikzpicture}
        \node[inner sep=0pt] at (72mm,75mm){\includegraphics[height=48mm]{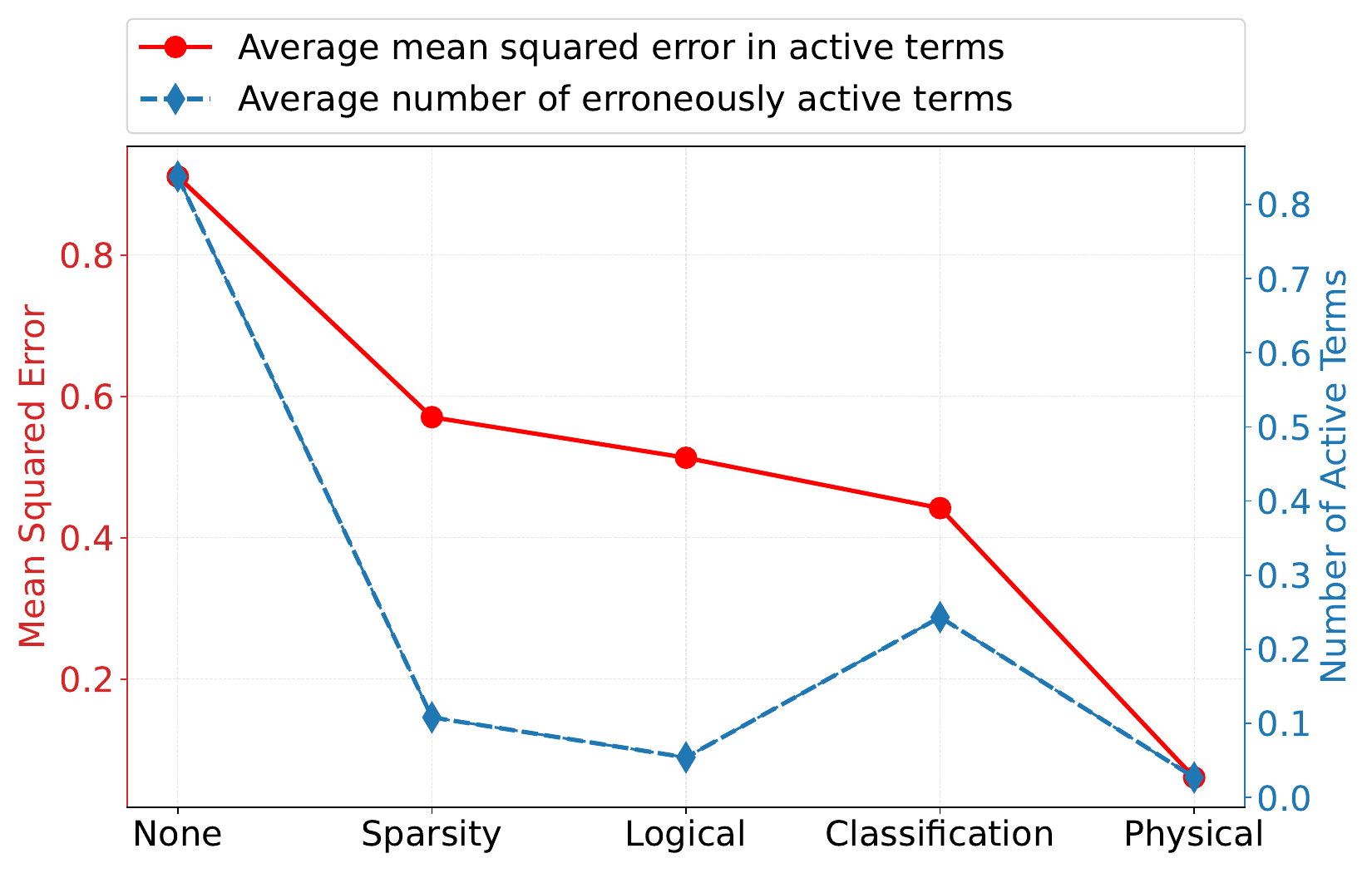}};
    \end{tikzpicture}
    \end{center}
    \caption{Representative latent spaces learned using progressively stronger constraints during training data generation, ranging from no constraints (None) to the full set of physics-based constraints (Physics). All models use an 11-dimensional latent space. As progressively stronger physical structure is incorporated into the training data, the learned representations more accurately reconstruct the benchmark PDEs, reducing both the coefficient error of active terms and the number of erroneously active terms.}
    \label{fig:ablation_test}
\end{figure}

To this end, separate GVAEs were trained using datasets generated from the nested hypothesis spaces \(\mathcal{X}, \mathcal{X}_{spr}, \mathcal{X}_{log}, \mathcal{X}_{fam}\) and \( \mathcal{X}_{phy}\). Figure~\ref{fig:ablation_test} reports the average reconstruction error over the 37 benchmark PDEs listed in Table~\ref{tbl:canonical_inc}, together with the average number of erroneously active terms. Although progressively removing the proposed principles substantially enlarges the admissible hypothesis space, all datasets were generated with approximately the same number of samples (\(\approx 120,000\)), ensuring that differences in performance can be attributed to the imposed inductive bias rather than dataset size.

Figure~\ref{fig:ablation_test} shows that when no inductive principles are enforced (i.e., the unrestricted hypothesis space \(\mathcal{X}\)), the average reconstruction error across the 37 benchmark equations exceeds 0.9 (solid red line with circle nodes), while approximately 0.8 additional terms are incorrectly activated in each recovered equation (dashed blue line with diamond nodes). This is expected because the unrestricted hypothesis space contains substantially greater structural variability than can be represented by an \(11\)-dimensional latent space. Although such a space could likely be represented by a GVAE, doing so would require a considerably higher latent dimensionality.

As progressively stronger inductive principles are introduced, the average number of erroneously active terms decreases from \(0.85 \) to \(0.02\). The largest improvements occur after enforcing sparsity and logical dependencies, indicating that these principles eliminate many structurally implausible combinations of equation terms before learning begins. An increase is observed after introducing PDE family classification, and disappears once physical admissibility is enforced. Overall, the physically constrained hypothesis space \( \mathcal{X}_{phy}\)  produces the lowest structural reconstruction error.

The average MSE, computed only over the active coefficients, decreases almost monotonically as progressively richer scientific principles are incorporated into the training distribution. Together, these results demonstrate that the proposed inductive principles do more than reduce the size of the hypothesis space; they introduce statistical regularities that the GVAE can exploit to learn a more representative latent representation of admissible governing equations.

\subsection{Continuity Within and Across Equation Families}
The previous results demonstrate that the learned latent representation accurately reconstructs a broad collection of admissible partial differential equations and organizes them into distinct regions of the latent space. However, successful Bayesian inference requires more than accurate reconstruction. It also requires that distances in the latent space correspond to meaningful changes in the underlying hypotheses, such that nearby latent points represent similar governing equations. To investigate this property, we examine how the learned representation evolves both within individual PDE families and across different families. Establishing this geometric continuity provides evidence that the latent space forms a suitable foundation for probabilistic reasoning over competing hypotheses.

\begin{figure}
    \begin{tikzpicture}

        \node[inner sep=0pt] at (72mm,75mm){\includegraphics[height=48mm]{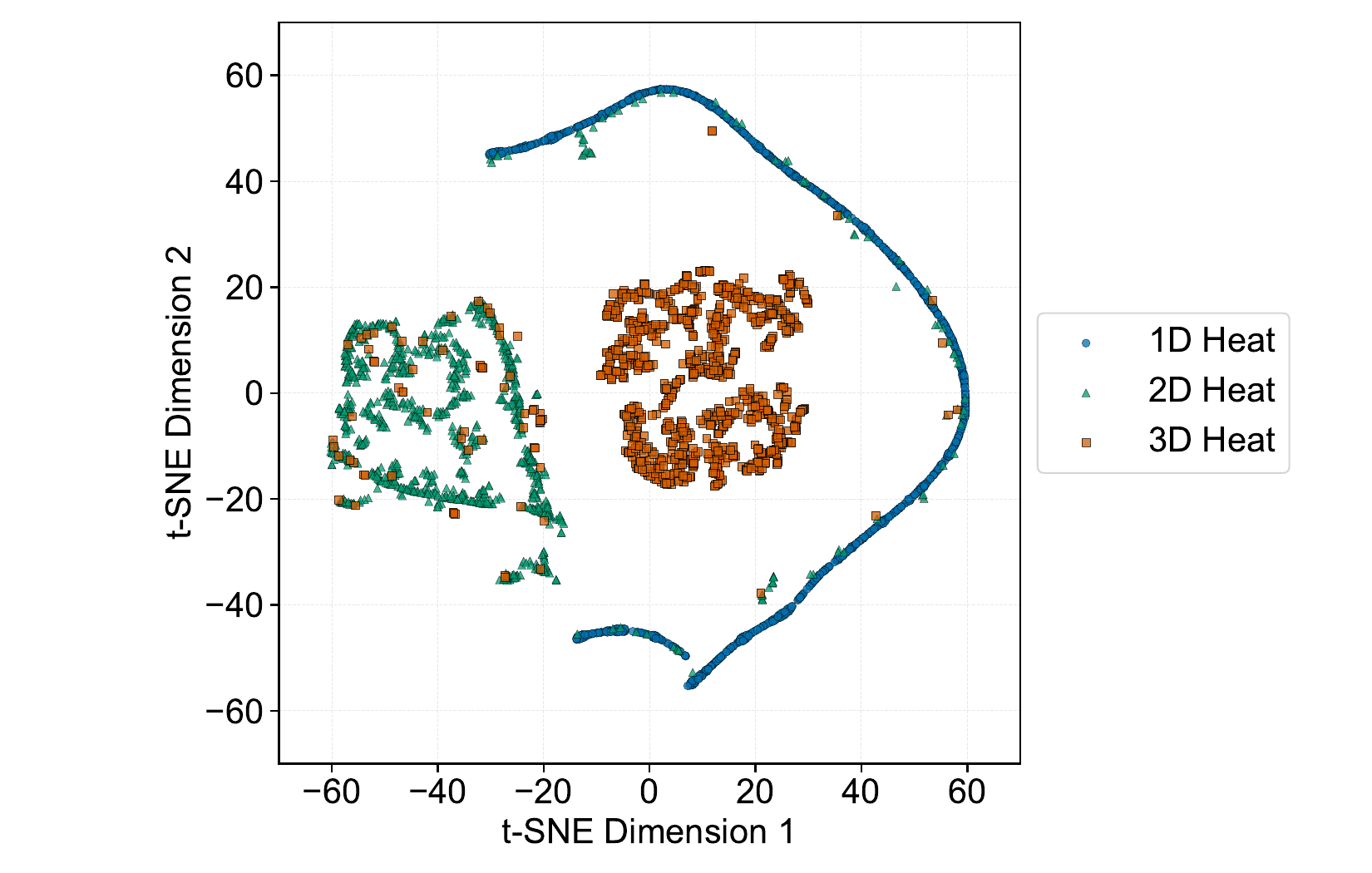}};
        \node[inner sep=0pt] at (72mm,25mm){\includegraphics[height=48mm]{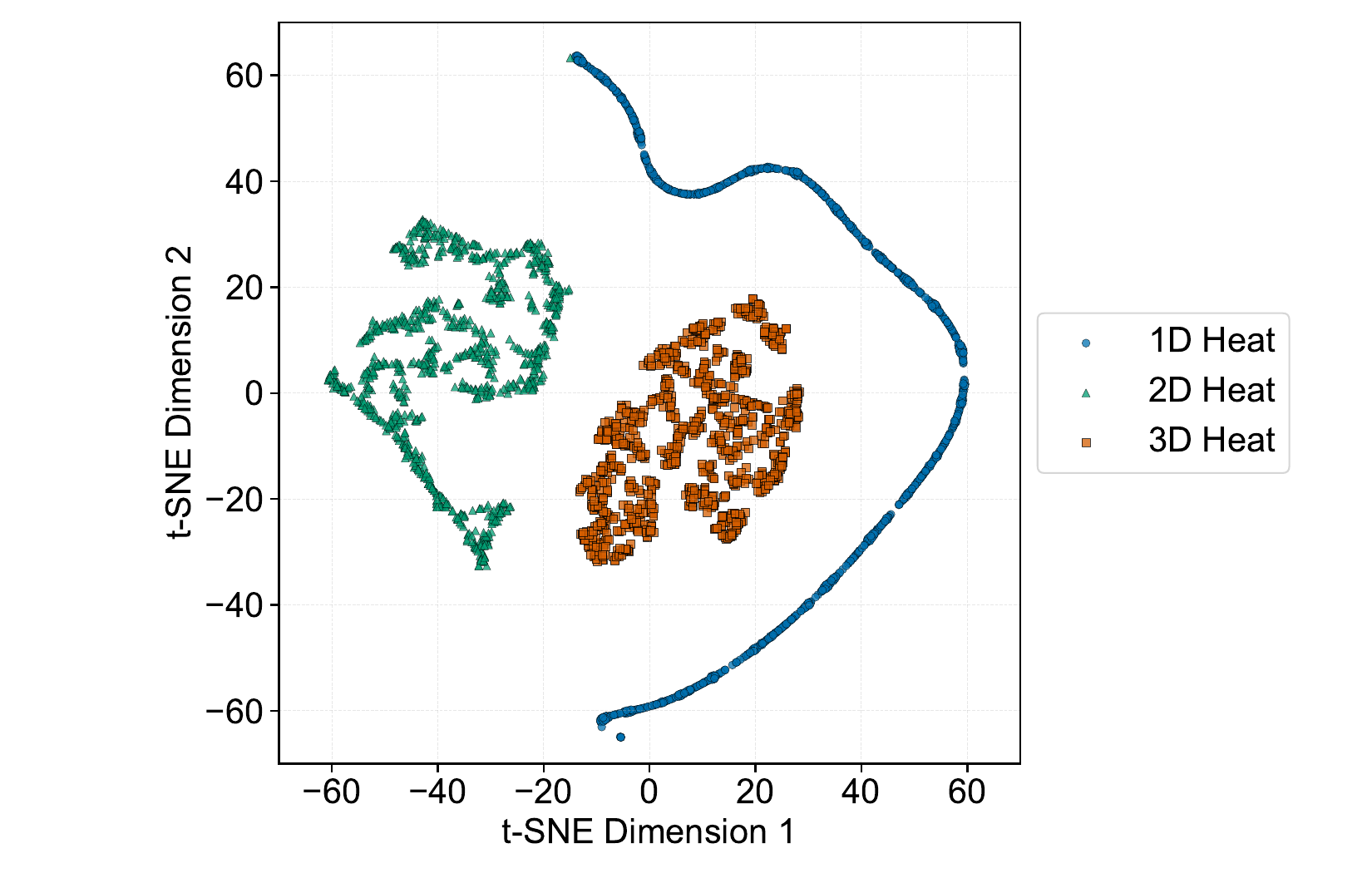}};

        \node[inner sep=0pt] at (-2mm,25mm){\includegraphics[height=48mm]{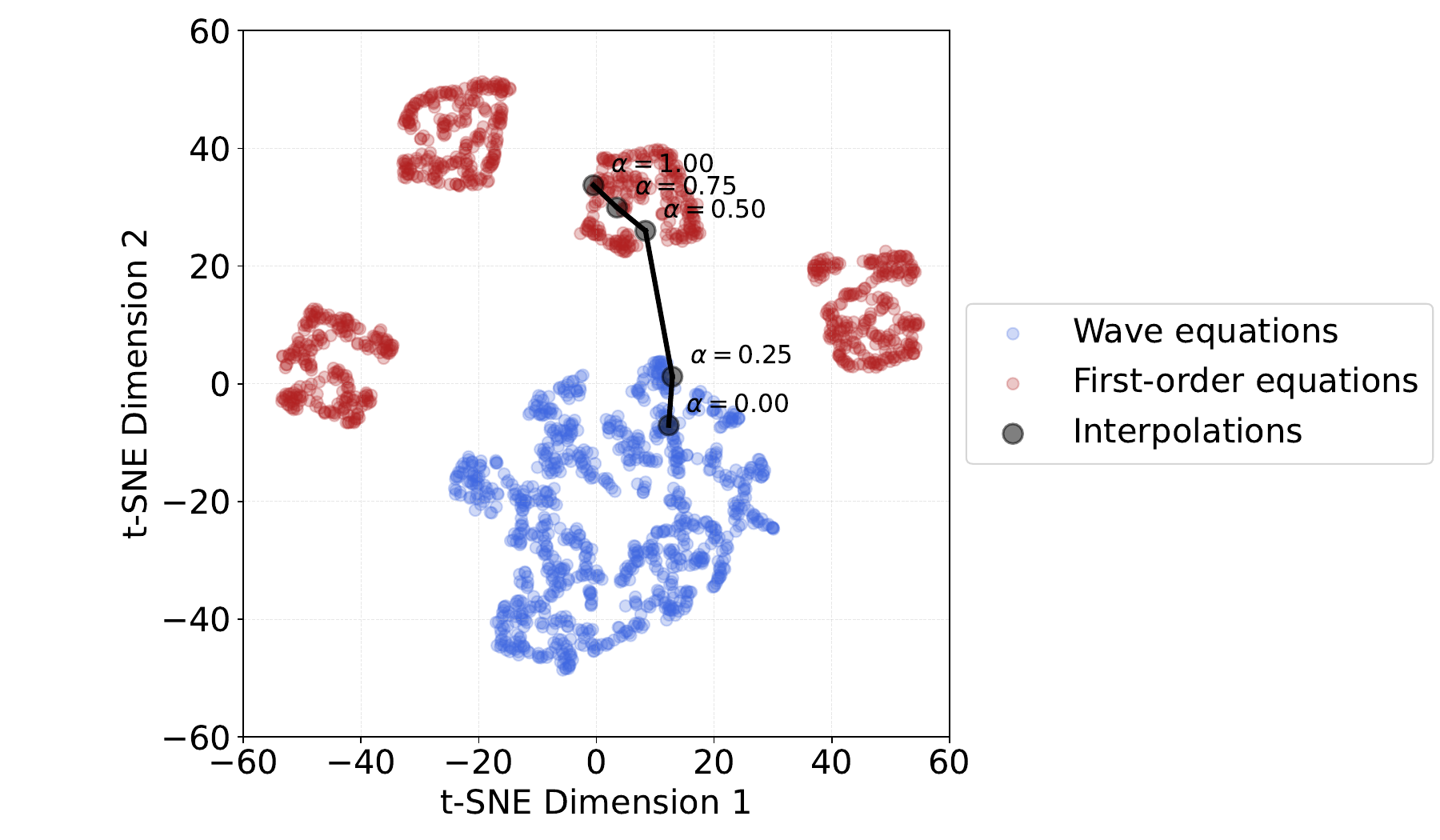}};
        \node[inner sep=0pt] at (-2mm,75mm){\includegraphics[height=48mm]{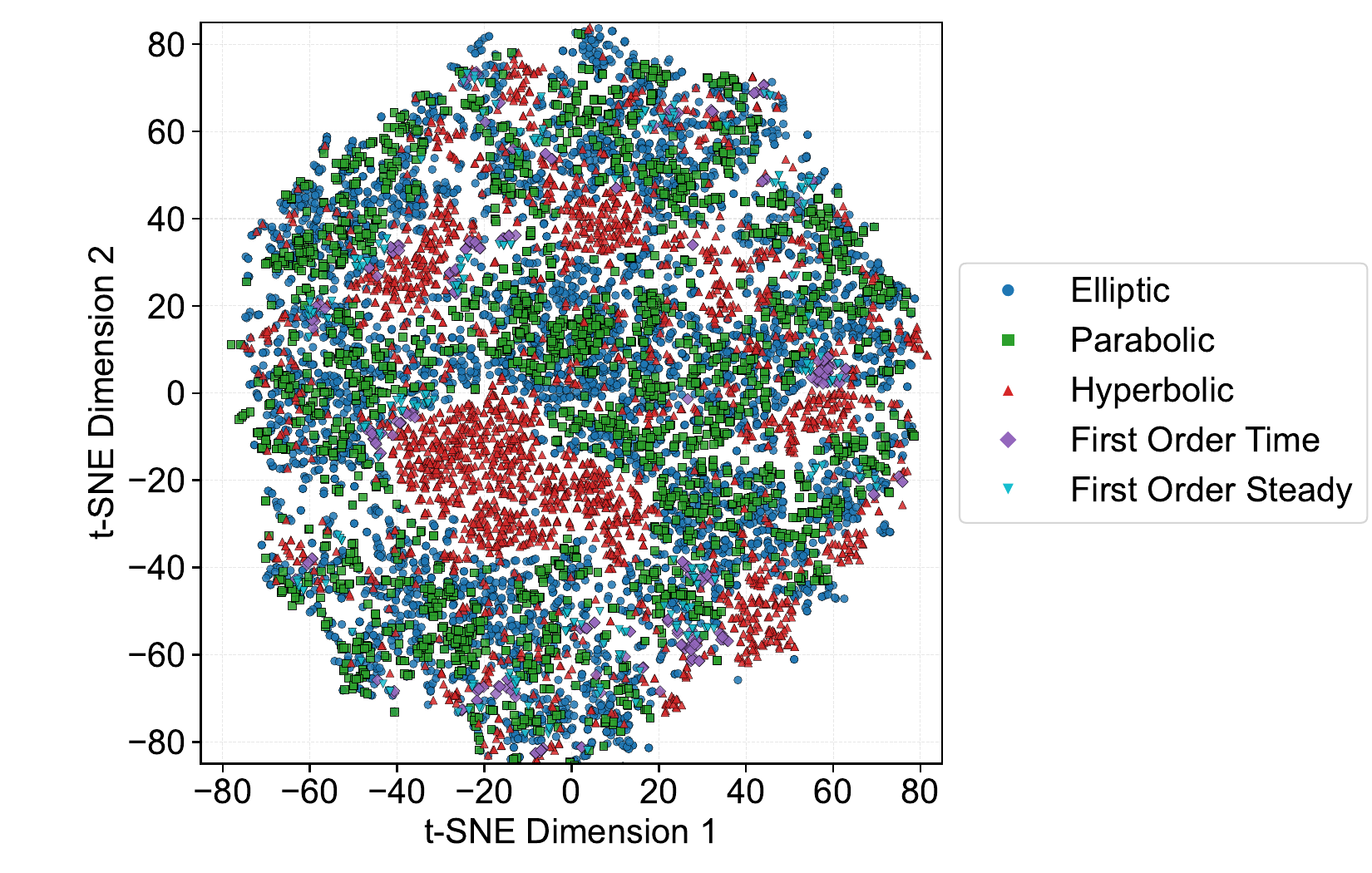}};

        \draw[line width=0.2mm, fill = white] (-38mm,96mm) --++ (4mm,0mm) --++ (0mm,4mm) --++ (-4mm,0mm) -- cycle;
        \node[anchor=center, align=center, text width = 5mm] at (-36mm,98mm) {\scriptsize \textbf{A}};

        \draw[line width=0.2mm, fill = white] (41mm,96mm) --++ (4mm,0mm) --++ (0mm,4mm) --++ (-4mm,0mm) -- cycle;
        \node[anchor=center, align=center, text width = 5mm] at (43mm,98mm) {\scriptsize \textbf{B}};

        \draw[line width=0.2mm, fill = white] (-38mm,46mm) --++ (4mm,0mm) --++ (0mm,4mm) --++ (-4mm,0mm) -- cycle;
        \node[anchor=center, align=center, text width = 5mm] at (-36mm,48mm) {\scriptsize \textbf{C}};

        \draw[line width=0.2mm, fill = white] (41mm,46mm) --++ (4mm,0mm) --++ (0mm,4mm) --++ (-4mm,0mm) -- cycle;
        \node[anchor=center, align=center, text width = 5mm] at (43mm,48mm) {\scriptsize \textbf{D}};
    \end{tikzpicture}
    \caption{Geometric organization and local smoothness of the learned latent space. A) Global organization according to PDE type (elliptic, parabolic, hyperbolic, first-order in time, and steady-state first-order). B) Organization according to PDE forms, revealing distinct but partially overlapping regions. C) Linear interpolation between a wave equation and a first-order time equation to verify if transitions between PDE families follow smooth trajectories in the latent space; the corresponding coefficients are reported in Table~\ref{tbl:interpolatebetweenforms}. D) Removing PDEs with coefficients below 0.1 from panel B exposes well-separated groups, indicating that changes between equation families arise through continuous evolution of the governing equation rather than abrupt jumps.}
    \label{fig:geomanalysis}
\end{figure}
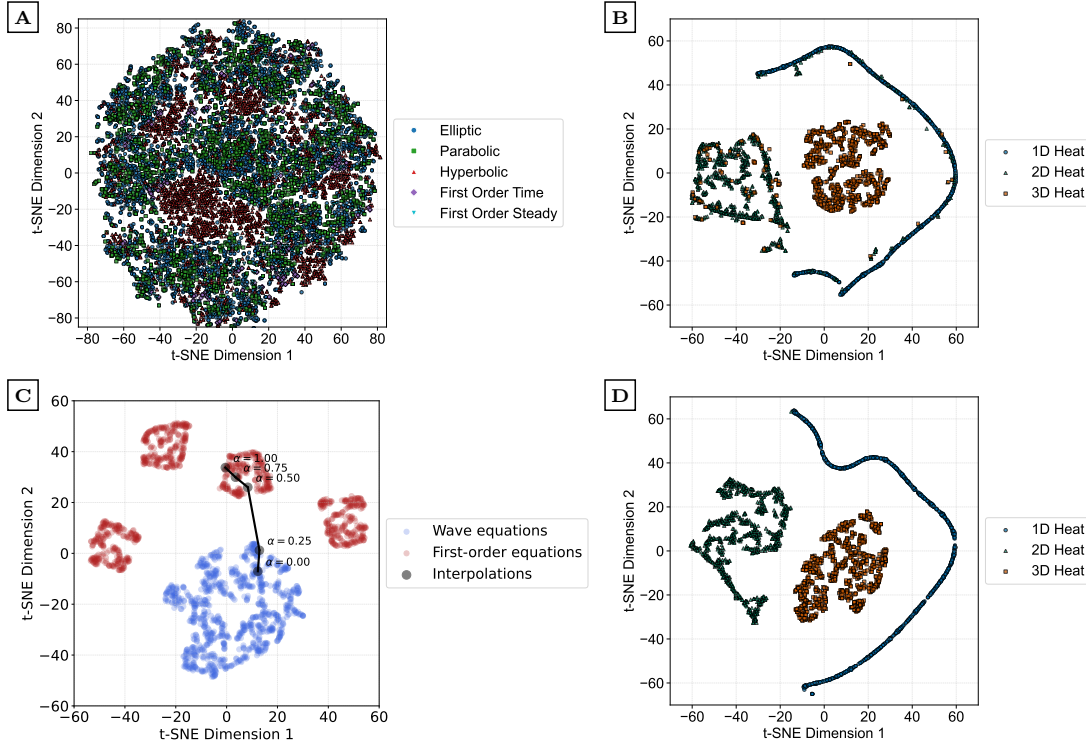

A reasonable expectation would be to expect that the GVAE would learn a latent space over the different PDE families are commonly used to classify different equations and that we enforced through Principle 3. However, this idea is only weakly supported through the two dimensional t-Sne plot presented in Panel A of Figure~\ref{fig:geomanalysis} where we can observe that the different PDE families do not form distinctive individual clusters. More specifically, the mean within-pattern distance for these five classes is \(3.106\) and the mean between-pattern distance is \(4.586\) resulting in a distance ration of \( 1.48\). This suggests that there is some structure according to these families; however, this amy not be the primary criteria that the GVAE used to distinguish between different PDE forms. 

Alternatively to the different PDE families, we plotted the two dimensional t-Sne plots for different parameterizations of the one, two, and three-dimensional heat equations in Panel B of Figure~\ref{fig:geomanalysis} though the blue circles, green triangles, and orange squared, respectively. Here we can observe a more stronger separation between the different PDEs, something that is quantitatively supported through their mean within-pattern distance of \(1.102\), and mean between-pattern distance of \(2.145\) for a distance ratio: \(1.950\). 

The overlap for some of the equations in Panel B of Figure~\ref{fig:geomanalysis} is because the latent space has learned a continuous representation between the different equations. Specifically, the three-dimensional heat equations that overlap the two dimensional heat equations have one diffusion term with a relatively low constant. This is validated by plotting the same t-Sne plot where we removed all the PDEs that have diffusion terms smaller than 0.1 in Panel D of Figure~\ref{fig:geomanalysis}. This shows that there are no longer any PDEs overlapping those with a different form. This observation suggests that the learned latent representation is sensitive not only to the symbolic structure of the governing equation but also to the effective strength of its differential operators. Consequently, PDEs with nearly inactive operators are embedded close to lower-dimensional counterparts, reflecting their similar physical behavior rather than their exact algebraic form.

While the different PDE forms provide neat clusters in the t-Sne plot, a question that we might ask ourselves is how smooth to equations transform within a same group. As such, lets consider that we want to samples two Wave, Heat, and Laplace equations and perform a linear interpolation in the latent space within these pairs. Let \(\mathbf{z}_1\) and \(\mathbf{z}_2\) be the latent coordinates for two randomly sampled equations of the same group, we can then conduct a linear interpolation in the latent space for a variable \(\alpha\in\left[0,1\right]\) as \(\mathbf{z}(\alpha)=(1-\alpha)\mathbf{z}_1+\alpha\mathbf{z}_2\). The reconstructed PDEs for a Wave, Heat, and Laplace equation have been presented in Column 3 of Table~\ref{tbl:interpolatewithinforms}, showing that there is an almost linear trend as one parametrization of an equations transition into another. 

\begin{table}
\setlength{\tabcolsep}{4.0pt}
\renewcommand{\arraystretch}{1}

\caption{Coefficients obtained by decoding equally spaced latent interpolations between different parameterizations of a Wave, Heat, and Laplace equations. The altering PDE forms and near constant latent space distance illustrate how changes are continuous and smooth.}
\centering
{\footnotesize 
\begin{tabular}{l l r l r r r}
\toprule
PDE & \( \alpha\) & \multicolumn{2}{l}{Reconstructed PDE} & Latent & Reconstruction & Latent \\
form &  & \multicolumn{2}{l}{} & distance \( (\%)\) & error \(\times 10^{2}\) & error \(\times 10^{2}\)\\
\midrule
Wave & 0 & \(0.696u_{tt}\) & \(= 0.594 u_{xx} + 0.346 u_{yy}\) & 0.0 & 2.90 & 0.88 \\
& 0.25 & \(0.591u_{tt}\) & \(= 0.618 u_{xx} + 0.388 u_{yy}\) & 25.0 & 2.03 & 0.71 \\
& 0.5 & \(0.485u_{tt}\) & \(= 0.642 u_{xx} + 0.431 u_{yy}\) & 50.0 & 5.04 & 1.12 \\
& 0.75 & \(0.354u_{tt}\) & \(= 0.664 u_{xx} + 0.477 u_{yy}\) & 75.0 & 4.06 & 1.36 \\
& 1 & \(0.240u_{tt}\) & \(= 0.686 u_{xx} + 0.523 u_{yy}\) & 100.0 & 6.04 & 1.46 \\

Heat & 0 & \(u_{t}\) & \(= 0.364 u_{xx}\) & 0.0 & 2.60 & 6.97 \\
& 0.25 & \(u_{t}\) & \(= 0.291 u_{xx}\) & 25.0 & 2.84 & 8.94 \\
& 0.5 & \(u_{t}\) & \(= 0.293 u_{xx} + 0.290 u_{yy}\) & 50.0 & 6.31 & 11.23 \\
& 0.75 & \(u_{t}\) & \(=0.291 u_{yy}\) & 75.0 & 6.36 & 7.97 \\
& 1 & \(u_{t}\) & \(=0.287 u_{yy}\) & 100.0 & 5.33 & 3.24 \\

Laplace & 0 & \(0\) & \(= 0.827 u_{xx} + 0.731 u_{yy}\) & 0.0 & 2.13 & 2.60 \\
& 0.25 & \(0\) & \(= 0.701 u_{xx} + 0.619 u_{yy}\) & 25.0 & 5.43 & 1.12 \\
& 0.5 & \(0\) & \(= 0.571 u_{xx} + 0.499 u_{yy}\) & 50.0 & 7.84 & 0.97 \\
& 0.75 & \(0\) & \(= 0.435 u_{xx} + 0.377 u_{yy}\) & 75.0 & 9.75 & 0.65 \\
& 1 & \(0\) & \(= 0.300 u_{xx} + 0.308 u_{yy}\) & 100.0 & 18.5 & 3.37 \\
\bottomrule
\end{tabular}
}
\label{tbl:interpolatewithinforms}
\end{table}

To study this further we also presented the reconstruction errors in Column 5 of Table~\ref{tbl:interpolatewithinforms} that have been calculated as \( ||\left( (1-\alpha)\mathbf{X}^{(1)}+\alpha\mathbf{X}^{(2)} \right) - d(\mathbf{z}(\alpha))||_2\) where \( d(\cdot)\) is the learned decoder, and \(\mathbf{X}^{(1)}\) and \(\mathbf{X}^{(2)}\) are the vector representations for the sampled equations. This shows that the linear interpolation in the latent space corresponds to a near linear interpolation in the actual vector representation of the PDEs as the reconstruction errors are mostly less than \(0.05\). Finally, we also present the latent errors in Column 6 of Table~\ref{tbl:interpolatewithinforms} that have been calculated as \( ||\mathbf{z}(\alpha)- e(d(\mathbf{z}(\alpha)))||_2/d_{avg}\), where \(e(\cdot)\) is the trained encoder and \(d_{avg}\) is the average distance between latent points so that we can interpret these errors as percentages. These relatively low errors show that that cycle consistency for the encoder and decor is relatively high, as most of them ar in the range of between \(0.5\%\) and \(5\%\). These show that the geometry of the latent space is able to meaningfully represent different equations of the same group.

\begin{table}
\setlength{\tabcolsep}{4.0pt}
\renewcommand{\arraystretch}{1}
\caption{Coefficients obtained by decoding equally spaced latent interpolations between a wave equation and a first-order time equation (Panel C of Figure~\ref{fig:geomanalysis}). For each latent point, the reconstructed coefficient is reported together with the decoder-estimated activation probability (shown in parentheses). The results illustrate that both coefficient values and symbolic equation structure evolve smoothly along the interpolation trajectory, demonstrating that transitions between PDE families correspond to continuous deformations in the learned latent representation.}
\centering
{\footnotesize
\begin{tabular}{l r l r r r r r r r r}
\toprule

 \(\alpha\) & \multicolumn{2}{l}{  Reconstructed PDE} &   \(u_t \) &   \(u_{tt} \) &  \( u_x \) &  \( u_y \) &  \( u_z \) &  \( u_{xx} \) &  \( u_{yy} \) &  \( u_{zz}\) \\
\midrule
 0    &   \(0.69u_{tt} \) &  \(   =0.33u_{xx}  + 0.31u_{yy}\)  & \makecell{ 0.72\\  (0.00)} & \makecell{ -0.69\\  (1.00)} & \makecell{ 0.04\\  (0.00)} & \makecell{ -0.01\\  (0.00)} & \makecell{ 0.02\\  (0.00)} & \makecell{ 0.33\\  (1.00)} & \makecell{ 0.31\\  (0.99)} & \makecell{ 0.42\\  (0.00)}\\
 0.25 &   \(0.41 u_{tt} \) &  \( = -0.12u_x  +0.29 u_{xx} +0.29 u_{yy} \) & \makecell{ 0.76\\  (0.00)} & \makecell{ -0.41\\  (1.00)} & \makecell{ -0.12\\  (0.99)} & \makecell{ 0.06\\  (0.30)} & \makecell{ -0.00\\  (0.00)} & \makecell{ 0.29\\  (0.99)} & \makecell{ 0.29\\  (0.75)} & \makecell{ 0.43\\  (0.00)}\\
 0.5  &   \( 0.19u_{tt} \) &  \( = -0.44u_x  +0.20u_y  +0.28u_{xx} \) & \makecell{ 0.27\\  (0.00)} & \makecell{ -0.19\\  (0.59)} & \makecell{ -0.44\\  (1.00)} & \makecell{ 0.20\\  (0.99)} & \makecell{ -0.01\\  (0.00)} & \makecell{ 0.28\\  (0.75)} & \makecell{ 0.27\\  (0.12)} & \makecell{ 0.43\\  (0.01)} \\
 0.75 &   \(0 \) &  \(=  0.72u_x  +0.41 u_y 
 \) & \makecell{ -0.77\\  (0.00)} & \makecell{ -0.13\\  (0.00)} & \makecell{ -0.72\\  (1.00)} & \makecell{ 0.41\\  (1.00)} & \makecell{ -0.01\\  (0.01)} & \makecell{ 0.25\\  (0.01)} & \makecell{ 0.25\\  (0.00)} & \makecell{ 0.48\\  (0.00)}\\
 1    &   \(u_t  \) &  \( = -0.90u_x  +0.59u_y  \) &\makecell{ -1.00\\  (0.97)} & \makecell{ -0.17\\  (0.00)} & \makecell{ -0.90\\  (1.00)} & \makecell{ 0.59\\  (1.00)} & \makecell{ -0.01\\  (0.01)} & \makecell{ 0.25\\  (0.00)} & \makecell{ 0.24\\  (0.01)} & \makecell{ 0.47\\  (0.00)} \\
\bottomrule
\end{tabular}
}
\label{tbl:interpolatebetweenforms}
\end{table}

An additional observation that we can from Table~\ref{tbl:interpolatewithinforms} is that for the Heat equation the diffusion term is gradually changing from \(u_{xx}\) to \(u_{yy}\). Specifically, we see that \(u_{xx}\) is turned of for \( \alpha>0.5\) and that \(u_{yy}\)is turned on for \( \alpha>0.25\) only turning into a two dimensional heat equation for \(\alpha=0.5\). This suggest that the learned latent space has a continuous and smooth transition between equations with a different form. To explore this further, we performed a linear latent space interpolation between a randomly selected Wave equation and first order time equation in Table~\ref{tbl:interpolatebetweenforms}. For the different values of \( \alpha\) we shown the reconstructed equations with the value for each term, and their binary activation probabilities in brackets. What we can observe here is that not only does the coefficient of a term change gradually, but also their activation probabilities. For example, the activation probability of \(u_{tt}\) is equal to 1 for \(\alpha=0\) which corresponding to a Wave equation, but the probability gradually decreases as it turn into a Wave equation for larger values of \(\alpha<0.5\). This transition has also been visualized in Panel C of Figure~\ref{fig:geomanalysis}. The black solid in this figure shows how equation continuously changes from a First-order equation, shown by the red circles, into a Wave equation as given by the blue circles. A final observation is that this figure shows four distinct clusters of equations associated with the first-order equations. This is due to the two advection terms (i.e., \(u_x\) and \(u_y\)) that have coefficients that can be positive and negative in sign. 
\section{Concluding Remarks}
We presented a systematic framework for learning continuous latent representations of admissible partial differential equations. Rather than introducing scientific knowledge through architectural constraints or modified optimization objectives, we embedded a progressively richer inductive bias directly into the training distribution. Specifically, we demonstrated how sparsity, logical dependencies, PDE family structure, and physical admissibility progressively organize the hypothesis space into a form that can be effectively learned by a gated variational autoencoder.

Experimental results showed that an 11-dimensional latent space is sufficient to accurately represent a broad collection of benchmark partial differential equations while exhibiting smooth geometric transitions both within and across equation families. Moreover, the ablation study demonstrated that progressively introducing scientific principles substantially improves the quality of the learned representation, supporting the central hypothesis that carefully designed training distributions can induce meaningful geometric structure in latent spaces. Together, these results suggest that distances within the learned latent space provide an interpretable measure of similarity between competing PDE hypotheses, making the representation a suitable foundation for inference over both equation structure and continuous coefficients.

Although the present work focuses on partial differential equations, the underlying approach is not restricted to this application. Many scientific hypothesis spaces consist of mixed discrete and continuous variables governed by domain-specific structural constraints. Extending the proposed framework to such domains will require identifying the corresponding scientific principles that define admissible hypotheses, rather than relying on the particular rules introduced for PDEs. From this perspective, the principal challenge is not the manifold learning algorithm itself, but the systematic construction of scientifically meaningful training distributions. Future work will therefore investigate Bayesian inference directly within the learned hypothesis manifold and explore analogous representations for other classes of scientific hypotheses. More broadly, this work suggests that scientific inductive bias can be embedded in the distribution of hypotheses rather than the architecture of the learning algorithm, providing a complementary perspective for developing probabilistic representations of scientific knowledge.

\bibliography{Bibliography}
\bibliographystyle{tmlr}


\end{document}